\documentclass{article}

\usepackage{graphicx}
\usepackage{booktabs}
\usepackage{amsmath}
\usepackage{tabularx}
\usepackage{makecell}
\usepackage{wrapfig}
\usepackage{textcomp}
\usepackage{tipa}
\usepackage{pgfplots}
\pgfplotsset{width=7cm,compat=1.9}
\usepackage{amssymb}
\usepackage{enumitem}
\usepackage{adjustbox}
\usepackage{sidecap}

\usepackage[preprint]{corl_2024} 

\title{\Huge Dreamitate:\\ \vspace{0.2cm} \Large Real-World Visuomotor Policy Learning via Video Generation}

\author{
  Junbang Liang$^{*1}$ \ \ Ruoshi Liu$^{*1}$ \ \ Ege Ozguroglu$^1$ \ \ Sruthi Sudhakar$^{1}$ \\ 
  \textbf{Achal Dave}$^{2}$ \ \ \textbf{Pavel Tokmakov}$^{2}$ \ \ \textbf{Shuran Song}$^{3}$ \ \ \textbf{Carl Vondrick}$^1$ \\
  \vspace{-0.2cm}
  \\ $^1$Columbia University \ \ $^2$Toyota Research Institute \ \ $^3$Stanford University \\
  \vspace{-0.2cm}
  \\ \href{https://dreamitate.cs.columbia.edu}{\url{dreamitate.cs.columbia.edu}}
  \vspace{-0.5cm}
}

\newcommand\blfootnote[1]{%
  \begingroup
  \renewcommand\thefootnote{}\footnote{#1}%
  \addtocounter{footnote}{-1}%
  \endgroup
}

\setlist{leftmargin=5.5mm}

\begin{document}
\maketitle
            
\blfootnote{$^*$ Equal contribution.}

\begin{figure}[h]
    \vspace{-2em}
    \centering
        \begin{adjustbox}{center, width=9em}
            \includegraphics{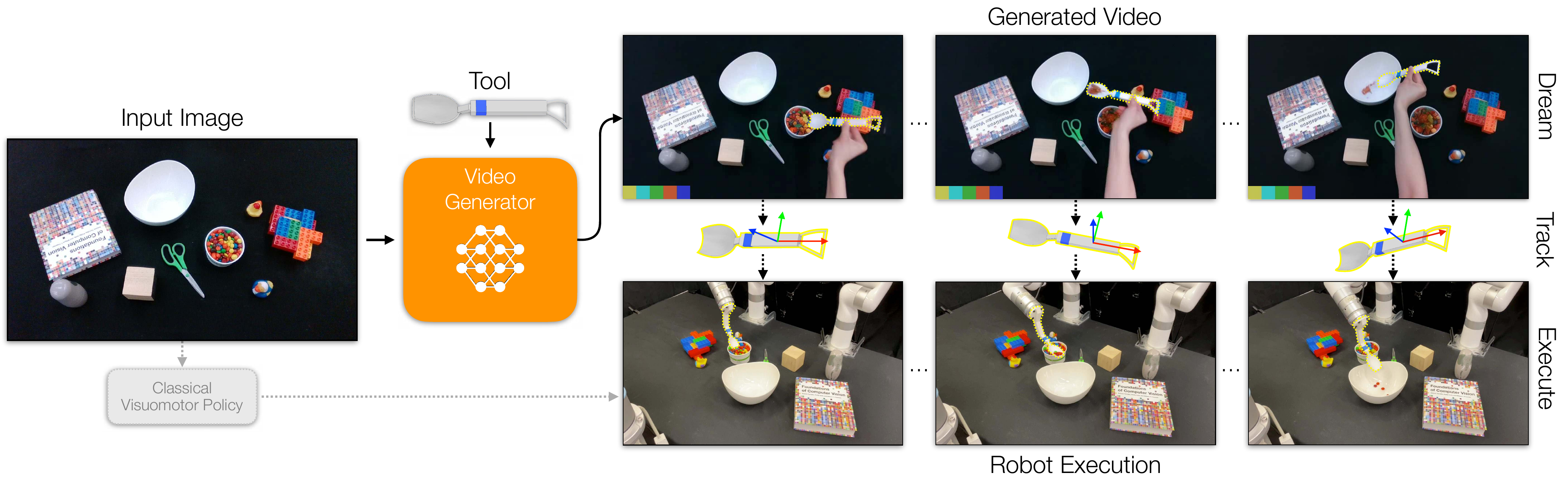}
        \end{adjustbox}
        \caption{\textbf{Real-World Visuomotor Policy Learning via Video Generation.} 
            Dreamitate is a visuomotor policy learning framework that fine-tunes a video generative model to synthesize videos (indicated by \includegraphics[width=1cm]{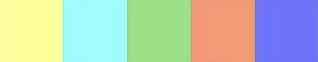}) of humans using tools to complete a task. The tool's trajectory in the generated video is tracked, and the robot executes this trajectory to accomplish the task in the real-world.}
            \label{fig:overview}
\end{figure}

\begin{abstract}
    A key challenge in manipulation is learning a policy that can robustly generalize to diverse visual environments. A promising mechanism for learning robust policies is to leverage video generative models that are pretrained on large-scale datasets of internet videos. In this paper, we propose a visuomotor policy learning framework that fine-tunes a video diffusion model on human demonstrations of a given task. At test time, we generate a video showing an execution of the task conditioned on images of a novel scene, and use this synthesized video to directly control the robot. Our key insight is that using common tools allows us to effortlessly bridge the embodiment gap between the human hand and the robot manipulator. We evaluate our approach on four tasks of increasing complexity and demonstrate that harnessing internet-scale generative models allows the learned policy to achieve a significantly higher degree of generalization than existing behavior cloning approaches.
\end{abstract}



\section{Introduction}




Learning visuomotor policies that decide actions from perception has been a longstanding challenge in robotics because they need to generalize to diverse physical situations \cite{khazatsky2024droid}. 
Recently, behavior cloning from human demonstrations~\cite{chi2024diffusion, zhao2023learning} has become the method of choice for many manipulation tasks, where visuomotor policy learning is formulated as a regression problem to map visual observations to actions supervised by human demonstrations. However, behavior cloning requires ground truth robot actions, which makes scaling to diverse situations challenging.


\looseness-1
We aim to learn generalizable visuomotor policies by leveraging video generation. Trained on large-scale datasets of internet videos, video generation provides a promising avenue for generalization because extensive priors from human behavior can be transferred into robot behavior. Many recent methods have also explored how to capitalize on these priors, such as synthesizing videos of human behavior~\cite{bharadhwaj2023towards} or synthesizing videos of robot behavior~\cite{du2023learning}.
However, while videos of people capture the diversity of behavior, generated human actions are difficult to transfer to robots due to the embodiment gap. While directly synthesizing videos of robots would be more realizable physically, the scale of available data is many orders of magnitude smaller than videos of humans, making the learned policy less robust in an in-the-wild environment.


\newcommand{\dictionaryentry}[4]{
    \noindent \textbf{#1} \quad | \textit{#2} | \texttt{#3} \\
    \begin{enumerate}[label=\textbf{\arabic*.},leftmargin=*]
        \item #4
    \end{enumerate}
}

How else can we leverage video generation for policy learning? In this work, we introduce Dreamitate,\footnote{Dream then imitate}
 a
visuomotor policy that controls robot behavior via conditional video generation. Our key insight is that the end-effector is the most important part of the embodiment for manipulation: its interaction with diverse objects is challenging to model explicitly and can benefit most from commonsense knowledge from video generation. The rest of robot embodiment can be solved through inverse kinematics.
Given a visual observation, our policy synthesizes a video of a person performing tasks with a tool. To translate into robot behaviors, we 3D track the tool in the synthesized video, and simply transfer the trajectory into explicit robot actions. Figure~\ref{fig:overview} shows an example of the actions created from this policy.

Our formulation offers several major advantages compared to traditional visuomotor policies:

\begin{itemize}
    \item \textbf{Generalizability.} Our underlying video generation model is pretrained on a massive amount of videos available on the internet, including a wide variety of human manipulation in all kinds of environments. By formulating policy learning as video model finetuning, the prior knowledge learned from pre-taining can be preserved, allowing our model to learn generalizable manipulation skills with the same amount of demonstration data.

    \item \textbf{Scalability.} Our finetuning videos are recordings of human demonstrations instead of teleoperation, making our dataset collection more scalable.

    \item \textbf{Interpretability.} During inference time, our video model predicts future execution plans in the form of videos before the actual robot executions are performed. Compared to a black-box end-to-end policy, our formulation offers an intermediate representation of policy that is interpretable to humans, a key feature for human-robot interaction applications.
\end{itemize}

We evaluated our approach across four real-world tasks, including bimanual manipulation, precise 3D manipulation, and long-horizon tasks, using a small number of expert human demonstrations. We found that the video model consistently outperformed the baseline behavior cloning model in generalizing to unseen scenarios. Additionally, we analyzed the model's performance when scaling down the training dataset and found that the video model maintained strong generalization performance even with fewer demonstrations. We will release the code and data for reproducing our results. Please refer to the supplementary material for the robot videos.

\section{Related Work}
\label{sec:related}

\textbf{Behavior Cloning.}
%
\looseness-1
Behavior cloning (BC) has emerged as a leading approach in manipulation tasks, learning policies from demonstrations through supervised learning. Early BC efforts involved end-to-end models mapping states to actions, but struggled with multimodal behaviors and high-precision tasks~\cite{pomerleau1988alvinn, zhang2018deep, florence2019selfsupervised}. Subsequently, Energy-Based Models (EBMs) were explored, predicting actions by minimizing energy in sequence optimization~\cite{lecun2006tutorial, du2020implicit, huang2023voxposer, florence2021implicit}. Recently, conditional generative models have shown promise in capturing multimodal demonstration behaviors, enhancing task success rates~\cite{chi2024diffusion, zhao2023learning, lee2024behavior}. Different from previous models, we integrate video prediction and 3D tracking to predict action prediction.

\textbf{Visual Pretraining for Policy Learning.}
Prior works have extensively explored various ways to pre-train the perception model in a visuomotor policy for learning more robust visual representation. One of the most popular pertaining objectives has been video prediction~\cite{finn2016unsupervised, sermanet2018timecontrastive, babaeizadeh2017stochastic, lee2018stochastic, Suris_2021_CVPR}. By learning to predict future based on the current observation, a model can learn the dynamic and causality of the world crucial for physical interaction. Contrastive learning~\cite{sermanet2018timecontrastive, srinivas2020curl, nair2022r3m, radford2021learning} as well as masked autoencoding~\cite{pmlr-v205-seo23a, pmlr-v205-radosavovic23a, radosavovic2023robot} are two very popular self-supervised learning objectives for learning visual representation for robotics. Besides, another line of work~\cite{ma2023liv, chen2021learning, escontrela2023video} studies learning a generalizable reward function through visual pertaining for reinforcement learning.

\textbf{Video Models for Decision-Making.}
More recently, in light of the fast progress of text-to-video generative models~\cite{videoworldsimulators2024, blattmann2023stable, blattmann2023align, zhang2023i2vgenxl, ho2022imagen}, the hope to leverage internet-scale video pertaining for robotics is rekindled~\cite{yang2024video}. A line of work uses video generative model as a world simulator~\cite{yang2023unisim, du2023learning, videoworldsimulators2024}, which can predict future video conditioned on an action. Another line of work uses video-language model for longer-horizon planning~\cite{du2023video, ajay2023compositional, black2023zero}. Different from prior works on this topic which typically use video prediction models as a world model or simulator in order to perform planning and decision making, Dreamitate uses video prediction model as part of a visuomotor policy and directly uses the predicted video for action prediction.

\vspace{-0.5em}



\begin{figure}[t]
\centering
\vspace{-1em}
\begin{adjustbox}{center, width=8em}
\includegraphics{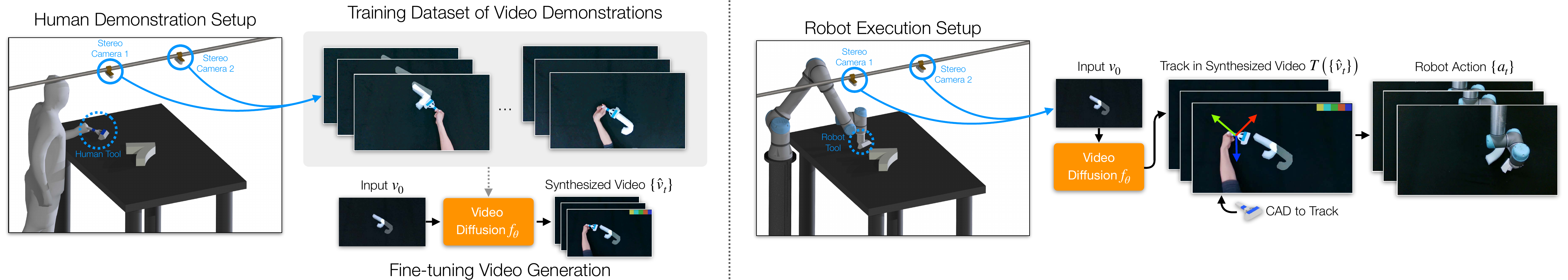}
\end{adjustbox}
\caption{\textbf{Method Overview.} For each task, (Step 1) we capture stereo camera recordings of human demonstrations of tool use from a top and a side view. (Step 2) The video model is fine-tuned to recreate these demonstrations from an initial scene image. (Step 3) In a new, unseen scenario, a stereo image pair is provided to the video model to generate (\includegraphics[width=1cm]{figures/generated_icon.png}) the manipulation. The tool's trajectory in the generated video is tracked and executed by the robot to complete the task.} \label{fig:method}
\end{figure}


\section{Approach} \label{approach}

\subsection{Overview}


Given a video frame of the scene $v_0$, our goal is to plan and execute robot actions $a_t \in \textrm{SE}(3)$ via video generation.
We approach this problem through the framework:
\begin{align}
     a_t = \mathcal{T}(\hat{v}_t) \quad \textrm{where} \quad \{\hat{v}_t\} = f_\theta(v_0)
\end{align}
where $f_\theta(\cdot)$ is a generative video model with learnable parameters $\theta$ and $\hat{v}_t$ is a synthesized frame. 
We train $f$ to generate videos of a human doing the task using a tool that we can track in 3D. $\mathcal{T}$ tracks the trajectory of the tool in the generated video $\hat{v}_t$, which we can directly use to control the end-effector state to perform manipulation tasks. Figure~\ref{fig:method} visualizes this framework.
This composition of synthesize-then-track facilitates cheap data collection without teleoperation. Moreover, it aligns with pretaining video data, which is composed of mostly videos showing human behavior.


\subsection{Video Generation} 

\begin{SCfigure}[1][t] 
  \vspace{-1em}
  \centering
  \includegraphics[width=0.58\textwidth]{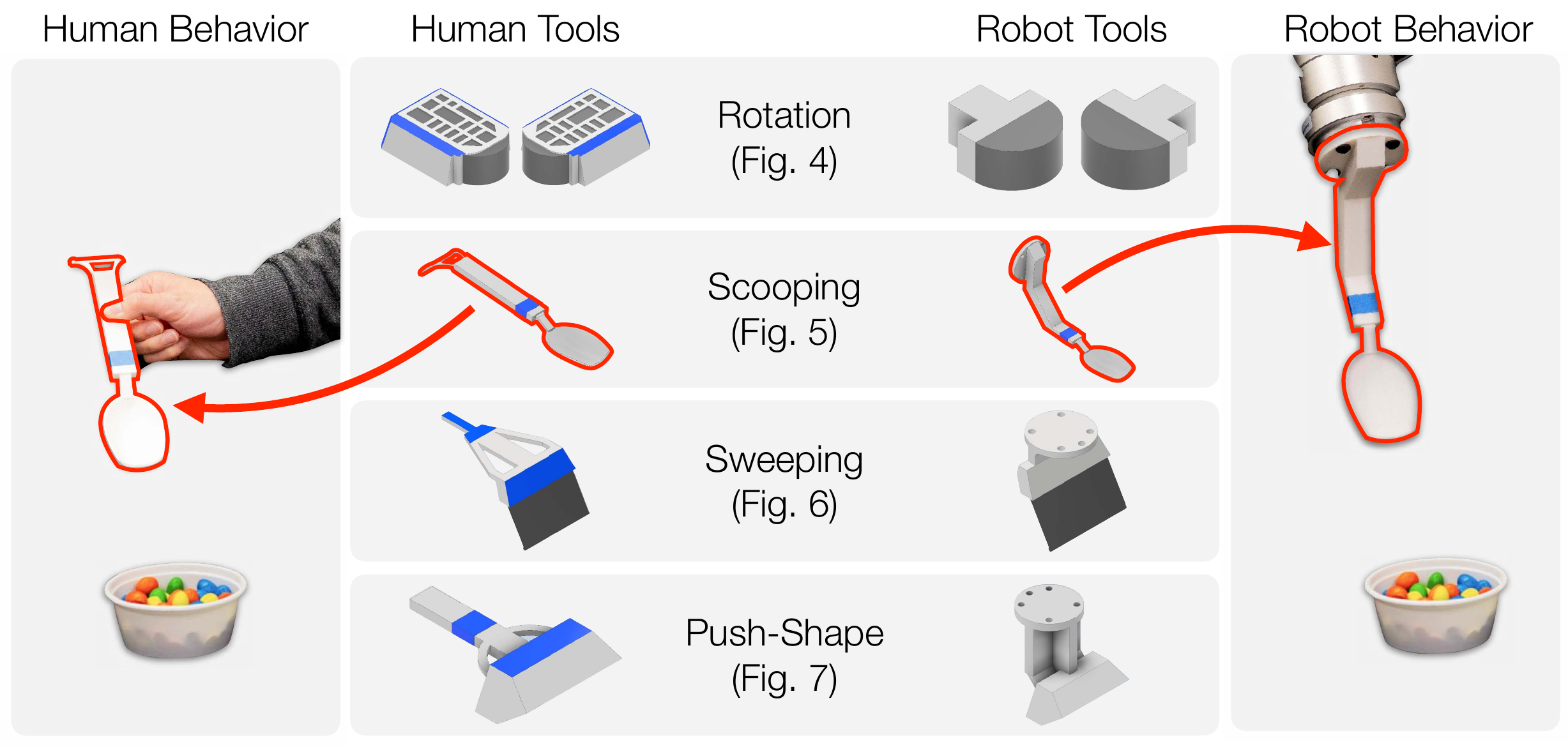} 
  \vspace{1em}
  \caption{\textbf{Aligning Human and Robot Behavior via Trackable Tools.} We use custom tools with known CAD models for human demonstrations, which facilitate precise 3D tracking, across four real-world manipulation tasks. We design unique tools (e.g., a spoon for scooping) to show the generalizability of our approach for varying tasks.}
  \label{fig:tools}
\end{SCfigure}

Starting with a video generator pre-trained on internet videos, we fine-tune it on a small video demonstration dataset in order to combine the priors from internet videos with behaviors from demonstration videos. To collect videos for fine-tuning, we build a tabletop setup with two calibrated cameras positioned 45 degrees apart for better visibility. We capture several pair of stereo videos $(\hat{v}^1, \hat{v}^2) \in \mathcal{V}$ of humans performing the task using a 3D printed tool (see Figure~\ref{fig:tools} for details).

We use these demonstration videos to fine-tune the video generative model. Since we will imitate the trajectory of the synthesized tool in the physical world, we generate stereo videos to make action possible in 3D.  Let $(\hat{v}_t^1, \hat{v}_t^2)$ be the predicted stereo video frames at time $t$ and $(v_t^1, v_t^2)$ be the ground truth stereo frames from demonstration videos. We optimize the video prediction objective:
\begin{align}
    \min_{\theta} \mathbb{E}_{v \in \mathcal{V}} \left[ \sum_{t=1}^T \left|\left|\left(\hat{v_t}^1 - v_t^1\right)\right|\right|_2 + \left|\left|\left(\hat{v_t}^2 - v_t^2\right)\right|\right|_2 \right] \quad \textrm{for} \quad \{\hat{v}_t\} = f_\theta(v_0)
\end{align}
We initialize $\theta$ to the pre-trained weights learned from large-scale internet video datasets (Stable Video Diffusion~\cite{blattmann2023stable}). We train a separate $f_\theta$ for each task (e.g. sweeping or scooping).

Following the implementation from~\cite{van2024generative}, the encoder and decoder are frozen such that only the spatial/temporal attention layers are fine-tuned. The per-frame image embedding input to the model is modified based on the viewing angle of the output frame to facilitate stereo video generation. This ensures that the first half of the frames generate the first view and the half generates the second view, using the initial image embedding from the respective view. 


At test time, we will be given a stereo image pair of the scene that we wish the robot to manipulate. We apply the fine-tuned video model $f(v_0)$ and obtain a generated stereo video frames $\hat{v}_t$ of the task being performed. 

\subsection{Track then Act}

The synthesized video frames $\{\hat{v}_t\}$ serve as the intermediate representation of our policy to obtain an action trajectory, $\{a_t\}$, that we then execute on the robot. Actions $a_t \in \mathrm{SE}(3)$ are represented as the 6D pose of the tool relative to the camera. By using a known CAD model, we can efficiently and accurately track the tool in each frame of the generated videos and therefore obtain precise locations of the tool. Each action $a_i$ in the action trajectory corresponds to one frame from the synthesized video. 
For 3D consistency, we use the generated stereo pair, along with the calibrated camera parameters to resolve 6D pose of the tool. For tracking the tool from an RGB image, we use Megapose~\cite{labbe2022megapose} and operate on $768 \times 448$ resolution video frames with camera intrinsic derived from the default values of the Intel Realsense camera. Finally, during execution, the customized tool is mounted on the robot arm and actions ${a_0,...,a_{T}}$ are executed by the robot arm. The entire data-collection, training, and fine-tuning pipeline are agent-independent. This is possible because we use tool's as manipulators and track known tools in generated videos, removing the need for learning any agent-specific action spaces.

\begin{figure}[t]
\vspace{-1em}
\centering
\includegraphics[width=\textwidth]{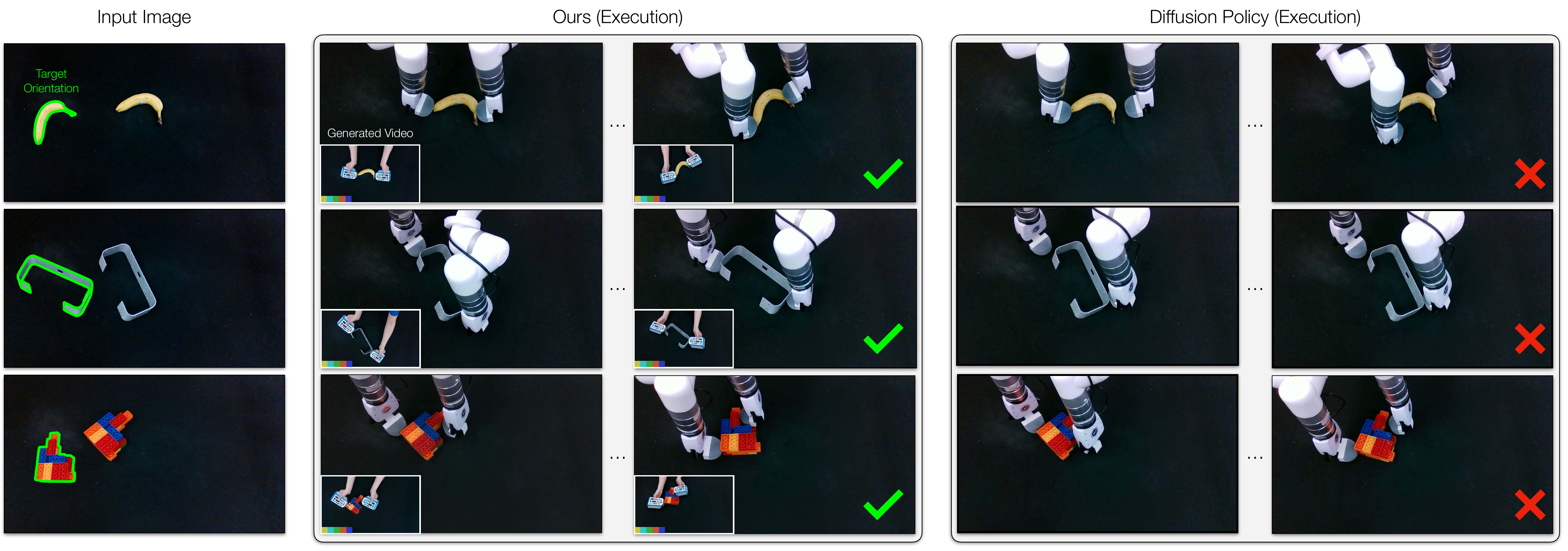}
\caption{\textbf{Rotation Qualitative Results.} For this task, we provide as input the image on the left for each row, with an example of a successful rotation overlaid. Our video generation-based approach succeeds, while Diffusion Policy fails to select stable grasping points on the object, causing it to slip during manipulation. Generated videos are indicated by \includegraphics[width=1cm]{figures/generated_icon.png}.} \label{fig:rotation_qualitative} \end{figure}

\section{Experiments}

In this section, we evaluate the performance of our video model in real-world robot manipulation tasks. The evaluation, summarized in Table~\ref{tab:expset}, encompasses four distinct tasks including object rotation, granular material scooping, table top sweeping and shape pushing (see Figure~\ref{fig:tools} for a visualization of the used tools). The details of each task are further explained in the corresponding sections below. For an assessment of our model's capabilities, we benchmark its performance against Diffusion Policy~\cite{chi2023diffusion} as a baseline.

\subsection{Experimental Setup}
\label{sec:experiment_setup}

To evaluate the model's generalizability, the data was collected by a human demonstrator at different locations for training and evaluation using the same camera setup, though the exact tabletop and lighting conditions varied. To further test the model's robustness in real-world scenarios, we ensured that the sets of objects used in training and evaluation are non-overlapping. To initialize tool tracking with Megapose~\cite{labbe2022megapose}, background subtraction and hand removal based on skin color were employed. In rare cases where this process failed, human corrections were applied to the bounding boxes (both for our method and for the baseline).

We used the same training data to train Diffusion Policy with stereo image input. The videos were preprocessed using Megapose for tool tracking, providing target trajectories for training. We used a pretrained ResNet-18 variant~\cite{chi2023diffusion} as the baseline, as the pretrained CLIP encoder~\cite{chi2024universal} variant showed lower performance in our tasks. We trained Diffusion Policy for 200 epochs to predict actions for the next 12 time steps, maintaining an open-loop system without subsampling the training data, similar to the video model.


\begin{table}[h]
    \centering
    \small
    \begin{tabularx}{\textwidth}{l>{\centering\arraybackslash}p{1.2in}>{\centering\arraybackslash}X>{\centering\arraybackslash}p{1.2in}>{\centering\arraybackslash}X}
        \toprule
         & Training Objects & Demonstrations & Test Objects & Test Trials\\
        \midrule
        Rotation & 31 & 371 & 10 & 40\\
        Scooping & 17 Bowls, 8 Particles & 368 & 8 Bowls, 4 Particles & 40\\
        Sweeping & 6 Particles & 356 & 6 Particles & 40\\
        Push-Shape & 26 Letters & 727 & 8 Shapes & 32\\
        \bottomrule
    \end{tabularx}
    \vspace{0.2cm}
    \caption{\textbf{Tasks Summary.} We detail the train and test setup for each task above. For each task, we use a distinct set of objects during testing than training. 
    }
    \label{tab:expset}
\end{table}

\vspace{-1em}
\subsection{Object Rotation Task}
\label{sec:rotation_experiment}

\textbf{Task.} We design a rotation task, shown in Figure~\ref{fig:rotation_qualitative}, to test our policy's ability to coordinate end-effectors and choose appropriate grasping points to manipulate real-world objects. In each training demonstration, we randomly place an object on the table, and use a gripper, shown in Figure~\ref{fig:tools} (top), to grasp the object at appropriate points and rotate it counterclockwise up to 45 degrees. 

\textbf{Setup.} We collect training data with 31 objects, with 14 real-world objects (such as boxes or a hammer) and 17 custom colored shapes made out of foam. 
We use 10 unseen real-world objects for evaluation.
We mark a trajectory as successful if the robot makes and maintains contact with the object throughout the rotation, and rotates it at least 25 degrees counterclockwise. 

\textbf{Results.} As shown in Table~\ref{tab:quantitative}, our policy  significantly outperforms Diffusion Policy (92.5\% vs. 55\%).
Fig.~\ref{fig:rotation_qualitative} illustrates cases where our policy succeeds while Diffusion Policy fails.
We observe that Diffusion Policy can fail to move the end-effector into contact with the object.
In more challenging cases, Diffusion Policy is prone to selecting unstable grasping points causing the object to slip during manipulation.
In contrast, our policy consistently makes contact with the object, with limited failures when selecting appropriate grasping points for particularly challenging shapes, such as a transparent bag with toys inside.

\begin{table}[h]
    \centering
    \small
    \begin{tabular}{cccccc}
        \toprule
        & Rotation & Scooping & Sweeping & \multicolumn{2}{c}{Push-Shape} \\
        \cmidrule(lr){2-2}
         \cmidrule(lr){3-3}
          \cmidrule(lr){4-4}
           \cmidrule(lr){5-6}
        Model & Success Rate & Success Rate & Success Rate & mIoU & Rot. Error \\
        \midrule
        Diffusion Policy~\cite{chi2024diffusion} & 22 / 40 & 22 / 40 & 5 / 40  & 0.550 & 48.2$^\circ$ \\
        \textbf{Ours} & \textbf{37 / 40} & \textbf{34 / 40} & \textbf{37 / 40} & \textbf{0.731} & \textbf{8.0$^\circ$} \\
        \bottomrule
    \end{tabular}
    \vspace{0.2cm}
    \caption{\textbf{Quantitative Results.} We compare our method to Diffusion Policy on four tasks quantitatively. We report success rates (successful trials / total trials) for rotation, scooping, and sweeping tasks. For Push-Shape, we report mean intersection-over-union (mIoU) and average rotation error (Rot. Error). Our approach performs well across all tasks, whereas Diffusion Policy shows worse performance overall and degrades in the more challenging sweeping and Push-Shape scenarios.}
    \vspace{-1em}
    \label{tab:quantitative}
\end{table}

\begin{figure}[t]
\vspace{-1em}
\centering
\includegraphics[width=\textwidth]{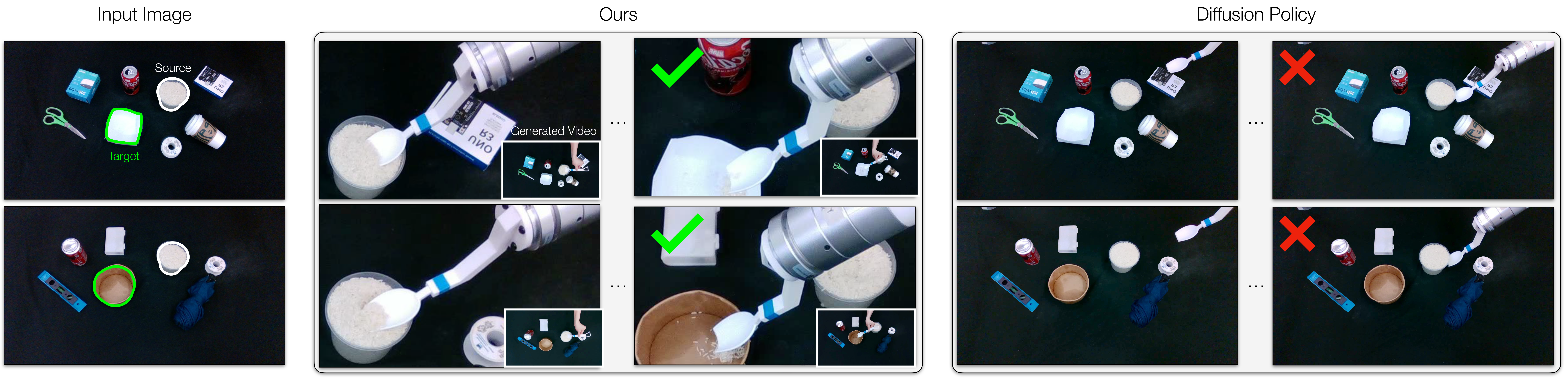}
\caption{\textbf{Scooping Qualitative Results.} We provide the input image on the left (without the overlays provided for illustration) to our model and Diffusion Policy, and show the output trajectory on the right. Our approach succeeds in most trials, while Diffusion Policy is often distracted by other objects in the scene, placing material in the wrong container and failing to scoop accurately. Generated videos are indicated by \includegraphics[width=1cm]{figures/generated_icon.png}.} \label{fig:scooping_qualitative} 
\vspace{-1em}
\end{figure}

\subsection{Granular Material Scooping Task}
\label{sec:scooping_experiment}

\textbf{Task.} 
This task, shown in Figure~\ref{fig:scooping_qualitative}, requires scooping granular material (e.g., beans) from a full container to an empty one while avoiding distractor objects.
This task requires the policy to perform precise manipulation of a scooping tool, identify the full and the empty containers, along with their precise locations, in arbitrary positions, and ignore distractors.

\textbf{Setup.} We collect demonstrations with 17 bowls and 8 colored beans with only 1 distraction object in the scene at a time.
At test time, we use an unseen set of 8 bowls, 4 new colored particles, and 5 distractor objects per trial sampled from a fixed set of 15 everyday objects. We mark a trajectory as successful if the robot transfers any particles to the empty bowl. 

\textbf{Results.} As shown in Table~\ref{tab:quantitative} (middle), our policy significantly outperforms Diffusion Policy (85\% vs. 55\%). This is a particularly challenging task for our approach, due to the small target (the scooper) for object tracking and video generation. 
This demonstrates that stereo video generation can accurately determine the object's pose to perform 3D manipulation. By comparison, Diffusion Policy is often distracted by objects in the scene, placing material in the wrong container, and failing more frequently in scooping due to misjudging the bowl's height on the table, as illustrated in Fig.~\ref{fig:scooping_qualitative}.

\begin{figure}[t]
\centering
\includegraphics[width=\textwidth]{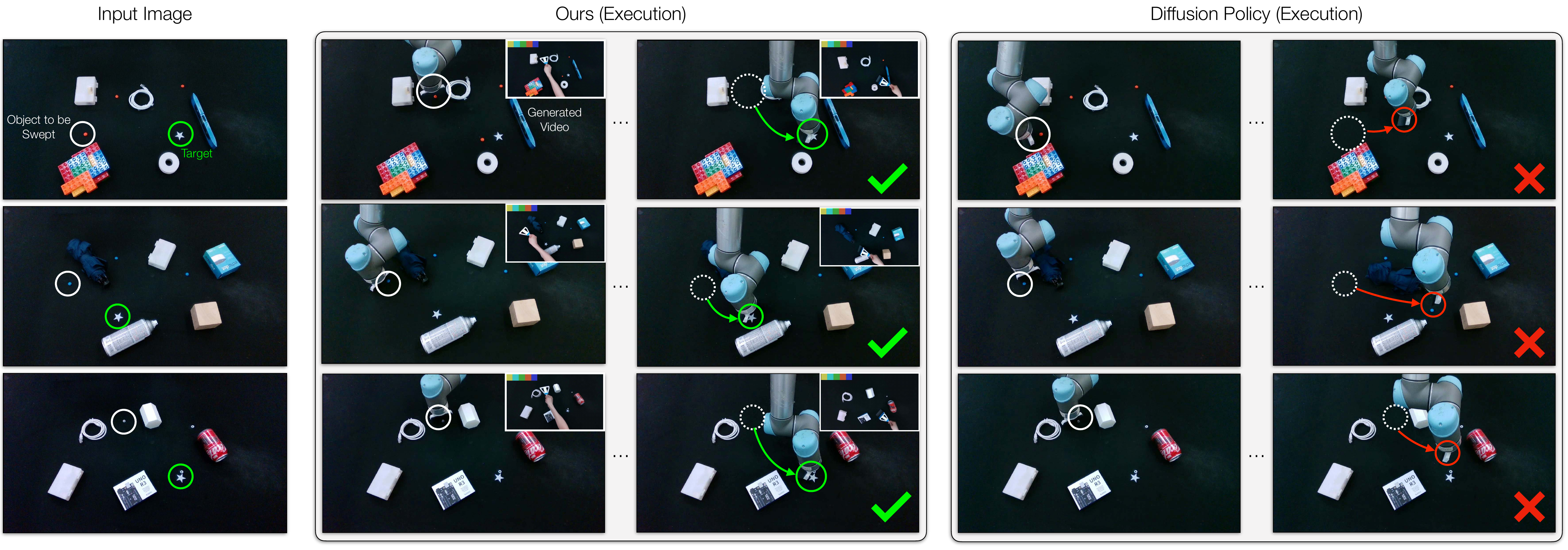}
\caption{\textbf{Sweeping Qualitative Results.} We provide the input image on the left (minus the white and green circles overlaid for illustration) to our model and Diffusion Policy, and show the output trajectories on the right. Our approach again achieves high success rate even on this challenging, multi-modal task, whereas Diffusion Policy generates trajectories that often collide with obstacles and fail to sweep to the target. Generated videos are indicated by \includegraphics[width=1cm]{figures/generated_icon.png}.} \label{fig:sweeping_qualitative} \end{figure}

\subsection{Table Top Sweeping Task}
\label{sec:sweeping_experiment}

\textbf{Task.}  The sweeping task, shown in Figure~\ref{fig:sweeping_qualitative}, requires policies to use a brush to sweep randomly placed particles to a target location marked by a randomly placed star, while avoiding obstacles.
This task is designed to test policy's ability to handle multi-modal distributions in the training data.

\textbf{Setup.} Training data includes 6 colored particles and 25 distraction objects, with 5 to 6 distractions at a time.
At test time, we use 6 new particles and 15 unseen distractions, with 2 to 4 beans and 5 distractions randomly placed on the table.
The training data contains some variations with multiple ways to achieve the goal, such as choosing which particle to sweep in the scene.
Due to the small size of the particles, we remove the final pooling layer from Diffusion Policy ResNet-18 encoder to provide a higher spatial resolution and improve baseline performance. We mark a trajectory as successful if any particle is transferred to be within 50 mm from the target. 

\textbf{Results.} In Table~\ref{tab:quantitative} (right), we observe that the video model maintains strong performance in this task with a 92.5\% success rate. In contrast, Diffusion Policy generates trajectories that often collide with obstacles and fail to sweep to the target (see Fig.~\ref{fig:sweeping_qualitative}), achieving only a 12.5\% success rate. This experiment demonstrates that the capitalizing on internet-scale pre-trained video generation models allows to better handle multi-modal demonstrations and achieve a much larger degree of generalization in this challenging scenario.

\subsection{Push-Shape Task (Long Horizon)}
\label{sec:pushing_experiment}

\begin{figure}[t]
\vspace{-1em}
\centering
\includegraphics[width=\textwidth]{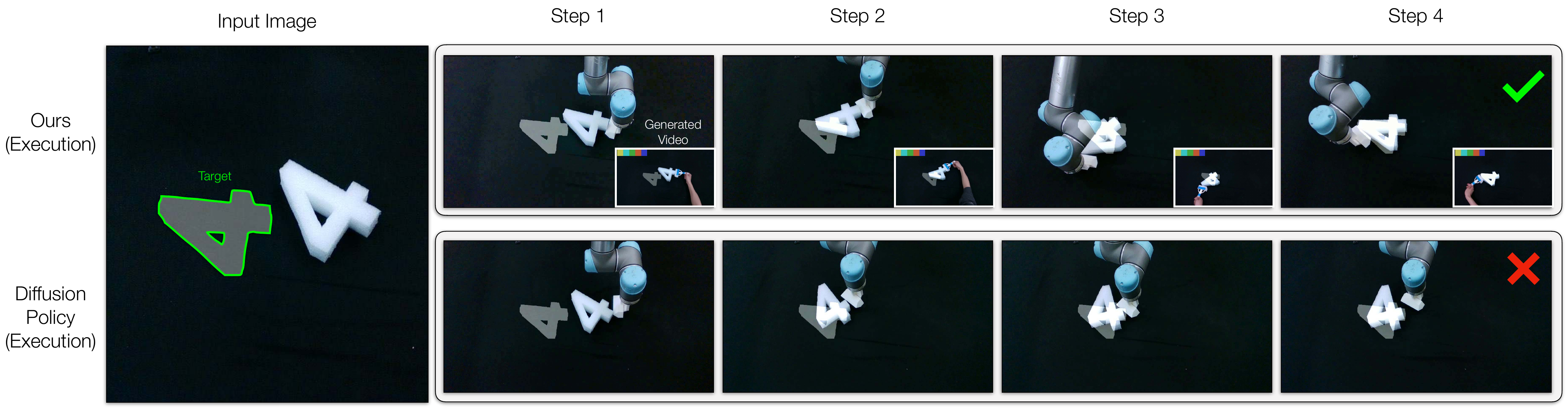}
\caption{\textbf{Push-Shape Qualitative Results.} We provide the image on the left, including the gray overlay indicating the target object position and orientation, to our model and Diffusion Policy, and show output trajectories on the right. Our method produces appropriate pushing actions to adjust the object's position. In contrast, Diffusion Policy fails to effectively rotate the object to match the mask. Generated videos (top row, bottom right) are indicated by \includegraphics[width=1cm]{figures/generated_icon.png}.} \label{fig:push_qualitative} \end{figure}

\textbf{Task.}
Push-Shape, shown in Figure~\ref{fig:push_qualitative}, is a challenging version of the long-horizon Push-T task~\cite{chi2024diffusion}: we place a foam shape on the table, and task the robot with pushing the object to a specified target goal mask (given as input to the policy) over consecutive steps.
This is a challenging task as it requires adjusting the position and orientation of the shape, requiring predicting the shape's movement, which depends on the contact between the object and the end-effector, as well as material properties of the table.
 
\textbf{Setup.} We train with a set of 26 foam objects each in the shape of a letter from the alphabet, and test on 8 unseen foam shapes (including digits and polygons).
As this is a challenging, multi-step task, we score the best of 4 rollouts for each trial, and report the mean intersect-over-union (mIoU) with the target mask as well as the average rotation error from the target.

%
\textbf{Results.}
As shown in Table~\ref{tab:quantitative}, over the 32 trials, the video model achieved a significantly higher mIoU of 0.731 compared to 0.550 mIoU from Diffusion Policy, and a rotation error of 8.0 degrees compared to 48.2 degrees from Diffusion Policy.
We find that Diffusion Policy tends to push the object to the target, but fails to effectively rotate the object to match the mask. 
By contrast, the video model produces appropriate pushing actions to adjust the object's position and orientation, thereby reducing the rotation error. Qualitative examples of the model rollouts are illustrated in Fig.~\ref{fig:push_qualitative}.

\begin{wrapfigure}[15]{r}{0.40\textwidth}
\centering
\begin{tikzpicture}[scale=0.8]
\begin{axis}[
    xlabel={Training Dataset Fraction},
    ylabel={Success Rate},
    xmin=0, xmax=10,
    ymin=0, ymax=1.1,
    xtick={1,5,9},
    xticklabels={1/3, 2/3, Full},
    ytick={0,0.2,0.4,0.6,0.8,1.0},
    legend style={at={(0.58,1.0)}, font=\small, scale=0.75}, 
    grid style=dashed,
]

\addplot[
    color={rgb:red,0;green,2.235;blue,3.705}, 
    mark=square,
    line width=1.5pt
    ]
    coordinates {
    (1,0.775)(5,0.75)(9,0.925)
    };
    \addlegendentry{Ours}

\addplot[
    color=orange, 
    mark=o,
    line width=1.5pt
    ]
    coordinates {
    (1,0.075)(5,0.25)(9,0.55)
    };
    \addlegendentry{Diffusion Policy}

\end{axis}
\end{tikzpicture}
\caption{\textbf{Number of Demonstrations.} Our approach maintains strong generalization performance with less training data compared to Diffusion Policy.} \label{fig:ablation}
\vspace{-3em}
\end{wrapfigure}
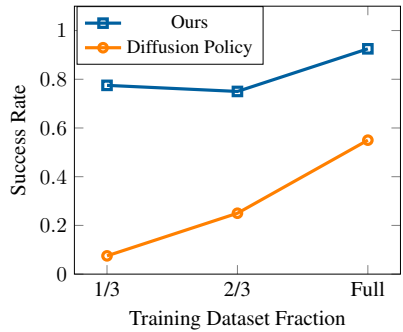

\newpage
\subsection{Performance Scaling Curve}
\label{sec:ablation}



We studied the impact of training set size on the generalization ability of our policy and compared it with Diffusion Policy using a rotation task. Both models performed well with the full dataset. When re-trained with two-thirds and one-third of the dataset and tested over 40 trials, the results depicted in Figure~\ref{fig:ablation} indicate that while Diffusion Policy's performance declines significantly with reduced data. In contrast, our model remains stable and maintains high success rates even with only one-third of the data. This highlights our model's superior generalization, which can be attributed to the extensive pre-training on Internet-scale video generation models.

\label{sec:result}


\section{Limitations}
\label{sec:limitations}

We highlight a few limitations and directions for future work. By visually tracking the tools used for manipulation, our implementation is limited to generating visually trackable robot actions. Although advances in video generation and object tracking models will improve accuracy, our approach can fail when the end-effector is heavily occluded. Additionally, reliance on rigid tools limits the applicability of our approach to the task requiring fine-grained control. Finally, video models have higher computational costs, making real-time closed-loop control infeasible, though this can be mitigated with recent advances ~\cite{luo2023lcm, li2023q} in accelerating video model inference.

\section{Conclusion}
\label{sec:conclusion}

In this work, we investigate how video generative models can be used to learn generalizable visuomotor policies. We propose to fine-tune a video diffusion model on human demonstrations to synthesize an execution plan in the form of videos at test-time. These synthesized execution videos are then directly used to control the robot. Our key insight is that using common tools allows us to easily bridge the embodiment gap between the human hand in the demonstrations and the robot manipulator in the real-world. Our experiments validate that capitalizing on Internet-scale video diffusion models allows our approach to achieve a much larger degree of generalization compared to previous behavior cloning methods like Diffusion Policy.


\clearpage
\acknowledgments{This research is based on work partially supported by the Toyota Research Institute and the NSF NRI Award \#2132519.}


\bibliography{example}  

\begin{thebibliography}{44}
\providecommand{\natexlab}[1]{#1}
\providecommand{\url}[1]{\texttt{#1}}
\expandafter\ifx\csname urlstyle\endcsname\relax
  \providecommand{\doi}[1]{doi: #1}\else
  \providecommand{\doi}{doi: \begingroup \urlstyle{rm}\Url}\fi

\bibitem[Khazatsky et~al.(2024)Khazatsky, Pertsch, Nair, Balakrishna, Dasari, Karamcheti, Nasiriany, Srirama, Chen, Ellis, Fagan, Hejna, Itkina, Lepert, Ma, Miller, Wu, Belkhale, Dass, Ha, Jain, Lee, Lee, Memmel, Park, Radosavovic, Wang, Zhan, Black, Chi, Hatch, Lin, Lu, Mercat, Rehman, Sanketi, Sharma, Simpson, Vuong, Walke, Wulfe, Xiao, Yang, Yavary, Zhao, Agia, Baijal, Castro, Chen, Chen, Chung, Drake, Foster, Gao, Herrera, Heo, Hsu, Hu, Jackson, Le, Li, Lin, Lin, Ma, Maddukuri, Mirchandani, Morton, Nguyen, O'Neill, Scalise, Seale, Son, Tian, Tran, Wang, Wu, Xie, Yang, Yin, Zhang, Bastani, Berseth, Bohg, Goldberg, Gupta, Gupta, Jayaraman, Lim, Malik, Martín-Martín, Ramamoorthy, Sadigh, Song, Wu, Yip, Zhu, Kollar, Levine, and Finn]{khazatsky2024droid}
A.~Khazatsky, K.~Pertsch, S.~Nair, A.~Balakrishna, S.~Dasari, S.~Karamcheti, S.~Nasiriany, M.~K. Srirama, L.~Y. Chen, K.~Ellis, P.~D. Fagan, J.~Hejna, M.~Itkina, M.~Lepert, Y.~J. Ma, P.~T. Miller, J.~Wu, S.~Belkhale, S.~Dass, H.~Ha, A.~Jain, A.~Lee, Y.~Lee, M.~Memmel, S.~Park, I.~Radosavovic, K.~Wang, A.~Zhan, K.~Black, C.~Chi, K.~B. Hatch, S.~Lin, J.~Lu, J.~Mercat, A.~Rehman, P.~R. Sanketi, A.~Sharma, C.~Simpson, Q.~Vuong, H.~R. Walke, B.~Wulfe, T.~Xiao, J.~H. Yang, A.~Yavary, T.~Z. Zhao, C.~Agia, R.~Baijal, M.~G. Castro, D.~Chen, Q.~Chen, T.~Chung, J.~Drake, E.~P. Foster, J.~Gao, D.~A. Herrera, M.~Heo, K.~Hsu, J.~Hu, D.~Jackson, C.~Le, Y.~Li, K.~Lin, R.~Lin, Z.~Ma, A.~Maddukuri, S.~Mirchandani, D.~Morton, T.~Nguyen, A.~O'Neill, R.~Scalise, D.~Seale, V.~Son, S.~Tian, E.~Tran, A.~E. Wang, Y.~Wu, A.~Xie, J.~Yang, P.~Yin, Y.~Zhang, O.~Bastani, G.~Berseth, J.~Bohg, K.~Goldberg, A.~Gupta, A.~Gupta, D.~Jayaraman, J.~J. Lim, J.~Malik, R.~Martín-Martín, S.~Ramamoorthy, D.~Sadigh, S.~Song, J.~Wu, M.~C. Yip, Y.~Zhu,
  T.~Kollar, S.~Levine, and C.~Finn.
\newblock Droid: A large-scale in-the-wild robot manipulation dataset, 2024.

\bibitem[Chi et~al.(2024)Chi, Xu, Feng, Cousineau, Du, Burchfiel, Tedrake, and Song]{chi2024diffusion}
C.~Chi, Z.~Xu, S.~Feng, E.~Cousineau, Y.~Du, B.~Burchfiel, R.~Tedrake, and S.~Song.
\newblock Diffusion policy: Visuomotor policy learning via action diffusion, 2024.

\bibitem[Zhao et~al.(2023)Zhao, Kumar, Levine, and Finn]{zhao2023learning}
T.~Z. Zhao, V.~Kumar, S.~Levine, and C.~Finn.
\newblock Learning fine-grained bimanual manipulation with low-cost hardware.
\newblock \emph{arXiv preprint arXiv:2304.13705}, 2023.

\bibitem[Bharadhwaj et~al.(2023)Bharadhwaj, Gupta, Kumar, and Tulsiani]{bharadhwaj2023towards}
H.~Bharadhwaj, A.~Gupta, V.~Kumar, and S.~Tulsiani.
\newblock Towards generalizable zero-shot manipulation via translating human interaction plans.
\newblock \emph{arXiv preprint arXiv:2312.00775}, 2023.

\bibitem[Du et~al.(2023)Du, Yang, Dai, Dai, Nachum, Tenenbaum, Schuurmans, and Abbeel]{du2023learning}
Y.~Du, M.~Yang, B.~Dai, H.~Dai, O.~Nachum, J.~B. Tenenbaum, D.~Schuurmans, and P.~Abbeel.
\newblock Learning universal policies via text-guided video generation, 2023.

\bibitem[Pomerleau(1988)]{pomerleau1988alvinn}
D.~A. Pomerleau.
\newblock Alvinn: An autonomous land vehicle in a neural network.
\newblock \emph{Advances in neural information processing systems}, 1, 1988.

\bibitem[Zhang et~al.(2018)Zhang, McCarthy, Jow, Lee, Chen, Goldberg, and Abbeel]{zhang2018deep}
T.~Zhang, Z.~McCarthy, O.~Jow, D.~Lee, X.~Chen, K.~Goldberg, and P.~Abbeel.
\newblock Deep imitation learning for complex manipulation tasks from virtual reality teleoperation.
\newblock In \emph{2018 IEEE international conference on robotics and automation (ICRA)}, pages 5628--5635. IEEE, 2018.

\bibitem[Florence et~al.(2019)Florence, Manuelli, and Tedrake]{florence2019selfsupervised}
P.~Florence, L.~Manuelli, and R.~Tedrake.
\newblock Self-supervised correspondence in visuomotor policy learning, 2019.

\bibitem[LeCun et~al.(2006)LeCun, Chopra, Hadsell, Ranzato, and Huang]{lecun2006tutorial}
Y.~LeCun, S.~Chopra, R.~Hadsell, M.~Ranzato, and F.~Huang.
\newblock A tutorial on energy-based learning.
\newblock \emph{Predicting structured data}, 1\penalty0 (0), 2006.

\bibitem[Du and Mordatch(2020)]{du2020implicit}
Y.~Du and I.~Mordatch.
\newblock Implicit generation and generalization in energy-based models, 2020.

\bibitem[Huang et~al.(2023)Huang, Wang, Zhang, Li, Wu, and Fei-Fei]{huang2023voxposer}
W.~Huang, C.~Wang, R.~Zhang, Y.~Li, J.~Wu, and L.~Fei-Fei.
\newblock Voxposer: Composable 3d value maps for robotic manipulation with language models, 2023.

\bibitem[Florence et~al.(2021)Florence, Lynch, Zeng, Ramirez, Wahid, Downs, Wong, Lee, Mordatch, and Tompson]{florence2021implicit}
P.~Florence, C.~Lynch, A.~Zeng, O.~Ramirez, A.~Wahid, L.~Downs, A.~Wong, J.~Lee, I.~Mordatch, and J.~Tompson.
\newblock Implicit behavioral cloning, 2021.

\bibitem[Lee et~al.(2024)Lee, Wang, Etukuru, Kim, Shafiullah, and Pinto]{lee2024behavior}
S.~Lee, Y.~Wang, H.~Etukuru, H.~J. Kim, N.~M.~M. Shafiullah, and L.~Pinto.
\newblock Behavior generation with latent actions.
\newblock \emph{arXiv preprint arXiv:2403.03181}, 2024.

\bibitem[Finn et~al.(2016)Finn, Goodfellow, and Levine]{finn2016unsupervised}
C.~Finn, I.~Goodfellow, and S.~Levine.
\newblock Unsupervised learning for physical interaction through video prediction, 2016.

\bibitem[Sermanet et~al.(2018)Sermanet, Lynch, Chebotar, Hsu, Jang, Schaal, and Levine]{sermanet2018timecontrastive}
P.~Sermanet, C.~Lynch, Y.~Chebotar, J.~Hsu, E.~Jang, S.~Schaal, and S.~Levine.
\newblock Time-contrastive networks: Self-supervised learning from video, 2018.

\bibitem[Babaeizadeh et~al.(2017)Babaeizadeh, Finn, Erhan, Campbell, and Levine]{babaeizadeh2017stochastic}
M.~Babaeizadeh, C.~Finn, D.~Erhan, R.~H. Campbell, and S.~Levine.
\newblock Stochastic variational video prediction.
\newblock \emph{arXiv preprint arXiv:1710.11252}, 2017.

\bibitem[Lee et~al.(2018)Lee, Zhang, Ebert, Abbeel, Finn, and Levine]{lee2018stochastic}
A.~X. Lee, R.~Zhang, F.~Ebert, P.~Abbeel, C.~Finn, and S.~Levine.
\newblock Stochastic adversarial video prediction, 2018.

\bibitem[Suris et~al.(2021)Suris, Liu, and Vondrick]{Suris_2021_CVPR}
D.~Suris, R.~Liu, and C.~Vondrick.
\newblock Learning the predictability of the future.
\newblock In \emph{Proceedings of the IEEE/CVF Conference on Computer Vision and Pattern Recognition (CVPR)}, pages 12607--12617, June 2021.

\bibitem[Srinivas et~al.(2020)Srinivas, Laskin, and Abbeel]{srinivas2020curl}
A.~Srinivas, M.~Laskin, and P.~Abbeel.
\newblock Curl: Contrastive unsupervised representations for reinforcement learning, 2020.

\bibitem[Nair et~al.(2022)Nair, Rajeswaran, Kumar, Finn, and Gupta]{nair2022r3m}
S.~Nair, A.~Rajeswaran, V.~Kumar, C.~Finn, and A.~Gupta.
\newblock R3m: A universal visual representation for robot manipulation, 2022.

\bibitem[Radford et~al.(2021)Radford, Kim, Hallacy, Ramesh, Goh, Agarwal, Sastry, Askell, Mishkin, Clark, Krueger, and Sutskever]{radford2021learning}
A.~Radford, J.~W. Kim, C.~Hallacy, A.~Ramesh, G.~Goh, S.~Agarwal, G.~Sastry, A.~Askell, P.~Mishkin, J.~Clark, G.~Krueger, and I.~Sutskever.
\newblock Learning transferable visual models from natural language supervision, 2021.

\bibitem[Seo et~al.(2023)Seo, Hafner, Liu, Liu, James, Lee, and Abbeel]{pmlr-v205-seo23a}
Y.~Seo, D.~Hafner, H.~Liu, F.~Liu, S.~James, K.~Lee, and P.~Abbeel.
\newblock Masked world models for visual control.
\newblock In K.~Liu, D.~Kulic, and J.~Ichnowski, editors, \emph{Proceedings of The 6th Conference on Robot Learning}, volume 205 of \emph{Proceedings of Machine Learning Research}, pages 1332--1344. PMLR, 14--18 Dec 2023.
\newblock URL \url{https://proceedings.mlr.press/v205/seo23a.html}.

\bibitem[Radosavovic et~al.(2023{\natexlab{a}})Radosavovic, Xiao, James, Abbeel, Malik, and Darrell]{pmlr-v205-radosavovic23a}
I.~Radosavovic, T.~Xiao, S.~James, P.~Abbeel, J.~Malik, and T.~Darrell.
\newblock Real-world robot learning with masked visual pre-training.
\newblock In K.~Liu, D.~Kulic, and J.~Ichnowski, editors, \emph{Proceedings of The 6th Conference on Robot Learning}, volume 205 of \emph{Proceedings of Machine Learning Research}, pages 416--426. PMLR, 14--18 Dec 2023{\natexlab{a}}.
\newblock URL \url{https://proceedings.mlr.press/v205/radosavovic23a.html}.

\bibitem[Radosavovic et~al.(2023{\natexlab{b}})Radosavovic, Shi, Fu, Goldberg, Darrell, and Malik]{radosavovic2023robot}
I.~Radosavovic, B.~Shi, L.~Fu, K.~Goldberg, T.~Darrell, and J.~Malik.
\newblock Robot learning with sensorimotor pre-training, 2023{\natexlab{b}}.

\bibitem[Ma et~al.(2023)Ma, Liang, Som, Kumar, Zhang, Bastani, and Jayaraman]{ma2023liv}
Y.~J. Ma, W.~Liang, V.~Som, V.~Kumar, A.~Zhang, O.~Bastani, and D.~Jayaraman.
\newblock Liv: Language-image representations and rewards for robotic control, 2023.

\bibitem[Chen et~al.(2021)Chen, Nair, and Finn]{chen2021learning}
A.~S. Chen, S.~Nair, and C.~Finn.
\newblock Learning generalizable robotic reward functions from "in-the-wild" human videos, 2021.

\bibitem[Escontrela et~al.(2023)Escontrela, Adeniji, Yan, Jain, Peng, Goldberg, Lee, Hafner, and Abbeel]{escontrela2023video}
A.~Escontrela, A.~Adeniji, W.~Yan, A.~Jain, X.~B. Peng, K.~Goldberg, Y.~Lee, D.~Hafner, and P.~Abbeel.
\newblock Video prediction models as rewards for reinforcement learning, 2023.

\bibitem[Brooks et~al.(2024)Brooks, Peebles, Holmes, DePue, Guo, Jing, Schnurr, Taylor, Luhman, Luhman, Ng, Wang, and Ramesh]{videoworldsimulators2024}
T.~Brooks, B.~Peebles, C.~Holmes, W.~DePue, Y.~Guo, L.~Jing, D.~Schnurr, J.~Taylor, T.~Luhman, E.~Luhman, C.~Ng, R.~Wang, and A.~Ramesh.
\newblock Video generation models as world simulators.
\newblock 2024.
\newblock URL \url{https://openai.com/research/video-generation-models-as-world-simulators}.

\bibitem[Blattmann et~al.(2023{\natexlab{a}})Blattmann, Dockhorn, Kulal, Mendelevitch, Kilian, Lorenz, Levi, English, Voleti, Letts, et~al.]{blattmann2023stable}
A.~Blattmann, T.~Dockhorn, S.~Kulal, D.~Mendelevitch, M.~Kilian, D.~Lorenz, Y.~Levi, Z.~English, V.~Voleti, A.~Letts, et~al.
\newblock Stable video diffusion: Scaling latent video diffusion models to large datasets.
\newblock \emph{arXiv preprint arXiv:2311.15127}, 2023{\natexlab{a}}.

\bibitem[Blattmann et~al.(2023{\natexlab{b}})Blattmann, Rombach, Ling, Dockhorn, Kim, Fidler, and Kreis]{blattmann2023align}
A.~Blattmann, R.~Rombach, H.~Ling, T.~Dockhorn, S.~W. Kim, S.~Fidler, and K.~Kreis.
\newblock Align your latents: High-resolution video synthesis with latent diffusion models, 2023{\natexlab{b}}.

\bibitem[Zhang et~al.(2023)Zhang, Wang, Zhang, Zhao, Yuan, Qin, Wang, Zhao, and Zhou]{zhang2023i2vgenxl}
S.~Zhang, J.~Wang, Y.~Zhang, K.~Zhao, H.~Yuan, Z.~Qin, X.~Wang, D.~Zhao, and J.~Zhou.
\newblock I2vgen-xl: High-quality image-to-video synthesis via cascaded diffusion models, 2023.

\bibitem[Ho et~al.(2022)Ho, Chan, Saharia, Whang, Gao, Gritsenko, Kingma, Poole, Norouzi, Fleet, and Salimans]{ho2022imagen}
J.~Ho, W.~Chan, C.~Saharia, J.~Whang, R.~Gao, A.~Gritsenko, D.~P. Kingma, B.~Poole, M.~Norouzi, D.~J. Fleet, and T.~Salimans.
\newblock Imagen video: High definition video generation with diffusion models, 2022.

\bibitem[Yang et~al.(2024)Yang, Walker, Parker-Holder, Du, Bruce, Barreto, Abbeel, and Schuurmans]{yang2024video}
S.~Yang, J.~Walker, J.~Parker-Holder, Y.~Du, J.~Bruce, A.~Barreto, P.~Abbeel, and D.~Schuurmans.
\newblock Video as the new language for real-world decision making, 2024.

\bibitem[Yang et~al.(2023)Yang, Chen, Wang, Manivasagam, Ma, Yang, and Urtasun]{yang2023unisim}
Z.~Yang, Y.~Chen, J.~Wang, S.~Manivasagam, W.-C. Ma, A.~J. Yang, and R.~Urtasun.
\newblock Unisim: A neural closed-loop sensor simulator, 2023.

\bibitem[Du et~al.(2023)Du, Yang, Florence, Xia, Wahid, Ichter, Sermanet, Yu, Abbeel, Tenenbaum, Kaelbling, Zeng, and Tompson]{du2023video}
Y.~Du, M.~Yang, P.~Florence, F.~Xia, A.~Wahid, B.~Ichter, P.~Sermanet, T.~Yu, P.~Abbeel, J.~B. Tenenbaum, L.~Kaelbling, A.~Zeng, and J.~Tompson.
\newblock Video language planning, 2023.

\bibitem[Ajay et~al.(2023)Ajay, Han, Du, Li, Gupta, Jaakkola, Tenenbaum, Kaelbling, Srivastava, and Agrawal]{ajay2023compositional}
A.~Ajay, S.~Han, Y.~Du, S.~Li, A.~Gupta, T.~Jaakkola, J.~Tenenbaum, L.~Kaelbling, A.~Srivastava, and P.~Agrawal.
\newblock Compositional foundation models for hierarchical planning, 2023.

\bibitem[Black et~al.(2023)Black, Nakamoto, Atreya, Walke, Finn, Kumar, and Levine]{black2023zero}
K.~Black, M.~Nakamoto, P.~Atreya, H.~Walke, C.~Finn, A.~Kumar, and S.~Levine.
\newblock Zero-shot robotic manipulation with pretrained image-editing diffusion models.
\newblock \emph{arXiv preprint arXiv:2310.10639}, 2023.

\bibitem[Van~Hoorick et~al.(2024)Van~Hoorick, Wu, Ozguroglu, Sargent, Liu, Tokmakov, Dave, Zheng, and Vondrick]{van2024generative}
B.~Van~Hoorick, R.~Wu, E.~Ozguroglu, K.~Sargent, R.~Liu, P.~Tokmakov, A.~Dave, C.~Zheng, and C.~Vondrick.
\newblock Generative camera dolly: Extreme monocular dynamic novel view synthesis.
\newblock \emph{arXiv preprint arXiv:2405.14868}, 2024.

\bibitem[Labb{\'e} et~al.(2022)Labb{\'e}, Manuelli, Mousavian, Tyree, Birchfield, Tremblay, Carpentier, Aubry, Fox, and Sivic]{labbe2022megapose}
Y.~Labb{\'e}, L.~Manuelli, A.~Mousavian, S.~Tyree, S.~Birchfield, J.~Tremblay, J.~Carpentier, M.~Aubry, D.~Fox, and J.~Sivic.
\newblock Megapose: 6d pose estimation of novel objects via render \& compare.
\newblock \emph{arXiv preprint arXiv:2212.06870}, 2022.

\bibitem[Chi et~al.(2023)Chi, Feng, Du, Xu, Cousineau, Burchfiel, and Song]{chi2023diffusion}
C.~Chi, S.~Feng, Y.~Du, Z.~Xu, E.~Cousineau, B.~Burchfiel, and S.~Song.
\newblock Diffusion policy: Visuomotor policy learning via action diffusion.
\newblock \emph{arXiv preprint arXiv:2303.04137}, 2023.

\bibitem[Chi et~al.(2024)Chi, Xu, Pan, Cousineau, Burchfiel, Feng, Tedrake, and Song]{chi2024universal}
C.~Chi, Z.~Xu, C.~Pan, E.~Cousineau, B.~Burchfiel, S.~Feng, R.~Tedrake, and S.~Song.
\newblock Universal manipulation interface: In-the-wild robot teaching without in-the-wild robots.
\newblock In \emph{Proceedings of Robotics: Science and Systems (RSS)}, 2024.

\bibitem[Luo et~al.(2023)Luo, Tan, Patil, Gu, von Platen, Passos, Huang, Li, and Zhao]{luo2023lcm}
S.~Luo, Y.~Tan, S.~Patil, D.~Gu, P.~von Platen, A.~Passos, L.~Huang, J.~Li, and H.~Zhao.
\newblock Lcm-lora: A universal stable-diffusion acceleration module.
\newblock \emph{arXiv preprint arXiv:2311.05556}, 2023.

\bibitem[Li et~al.(2023)Li, Liu, Lian, Yang, Dong, Kang, Zhang, and Keutzer]{li2023q}
X.~Li, Y.~Liu, L.~Lian, H.~Yang, Z.~Dong, D.~Kang, S.~Zhang, and K.~Keutzer.
\newblock Q-diffusion: Quantizing diffusion models.
\newblock In \emph{Proceedings of the IEEE/CVF International Conference on Computer Vision}, pages 17535--17545, 2023.

\bibitem[He et~al.(2016)He, Zhang, Ren, and Sun]{he2016deep}
K.~He, X.~Zhang, S.~Ren, and J.~Sun.
\newblock Deep residual learning for image recognition.
\newblock In \emph{Proceedings of the IEEE conference on computer vision and pattern recognition}, pages 770--778, 2016.

\end{thebibliography}

\newpage
\appendix

\section{Video Model Implementation}

\begin{figure}[h]
\centering
\includegraphics[width=0.85\textwidth]{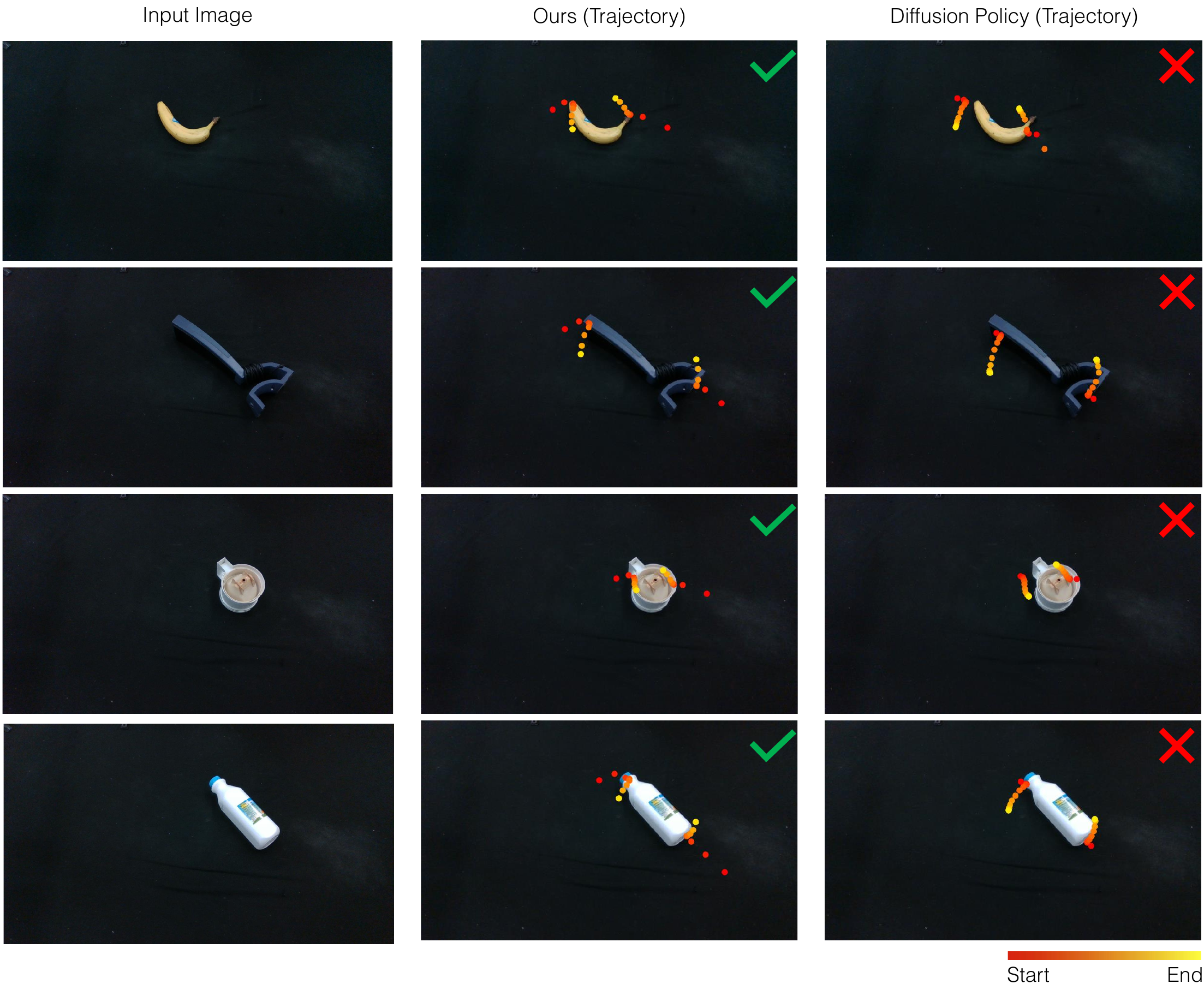}
\caption{\textbf{Additional Rotation Qualitative Results.} The trajectories of the end-effectors are projected onto the input image.} \label{fig:appendix_rotate_crop} \end{figure}

\begin{figure}[h]
\centering
\includegraphics[width=0.85\textwidth]{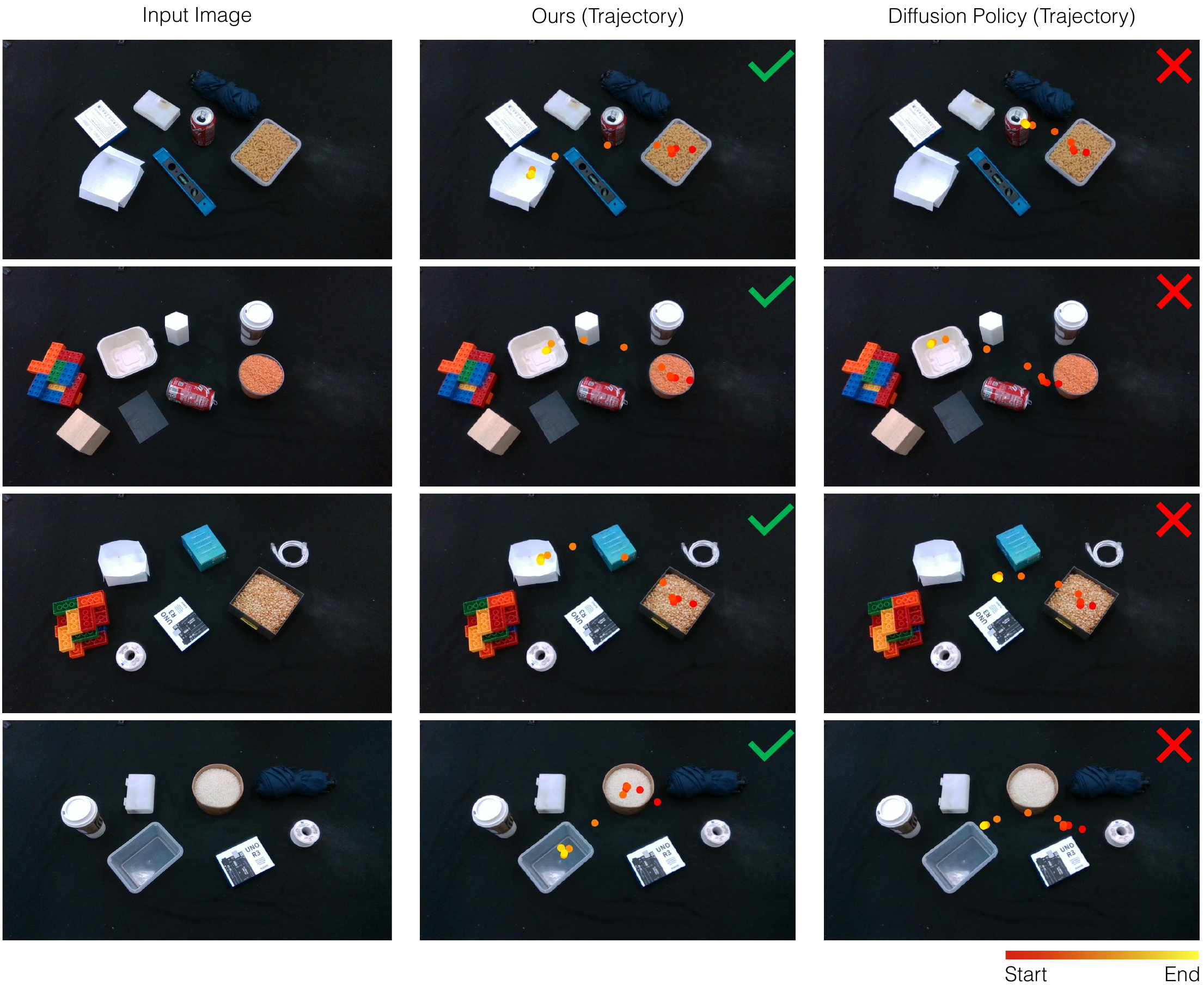}
\caption{\textbf{Additional Scooping Qualitative Results.} The trajectory of the end-effector is projected onto the input image.} \label{fig:appendix_scoop_crop} \end{figure}

\begin{figure}[h]
\centering
\includegraphics[width=0.85\textwidth]{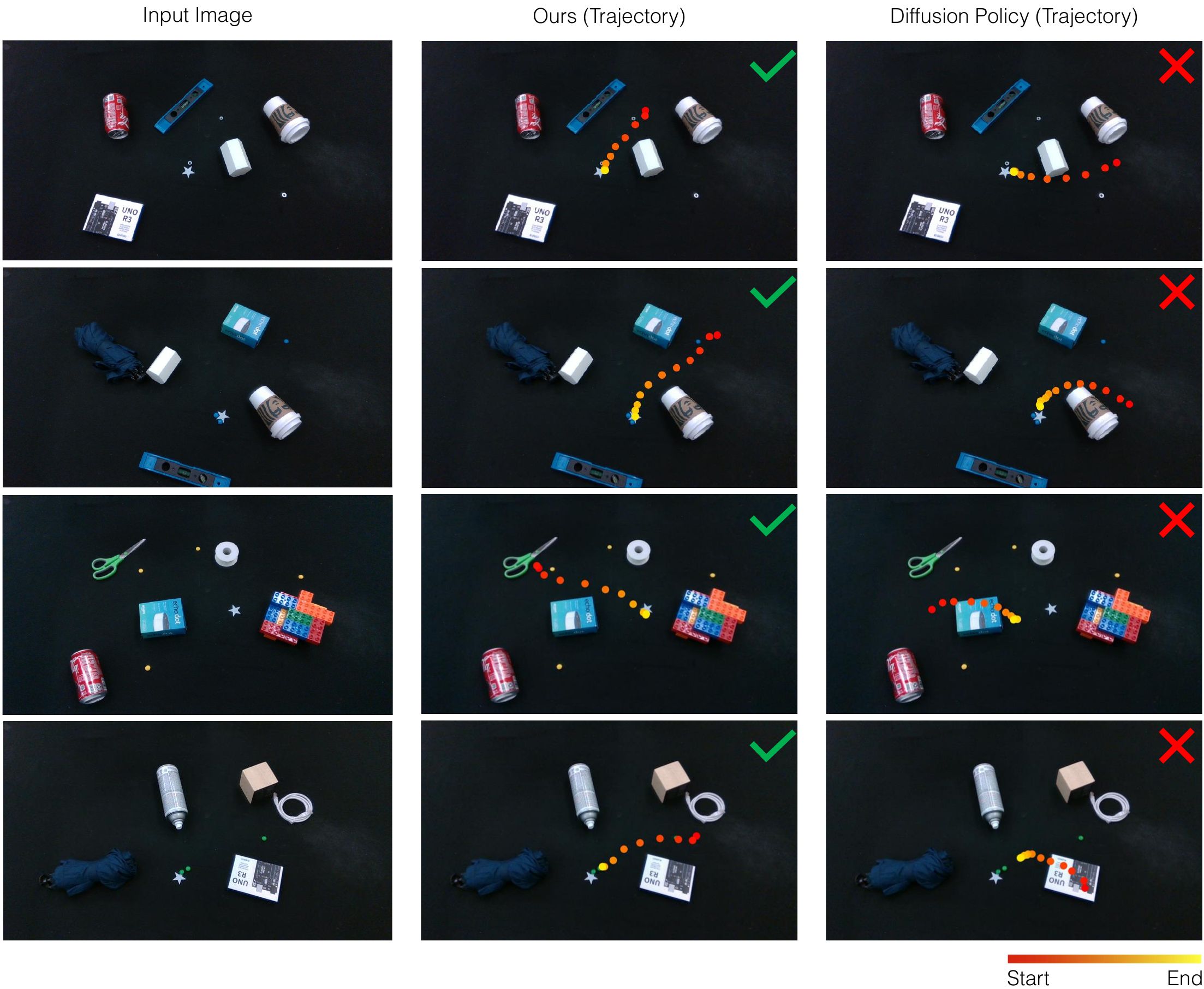}
\caption{\textbf{Additional Sweeping Qualitative Results.} The trajectory of the end-effector is projected onto the input image.} \label{fig:appendix_sweep_crop} \end{figure}

\begin{figure}[h]
\centering
\includegraphics[width=0.95\textwidth]{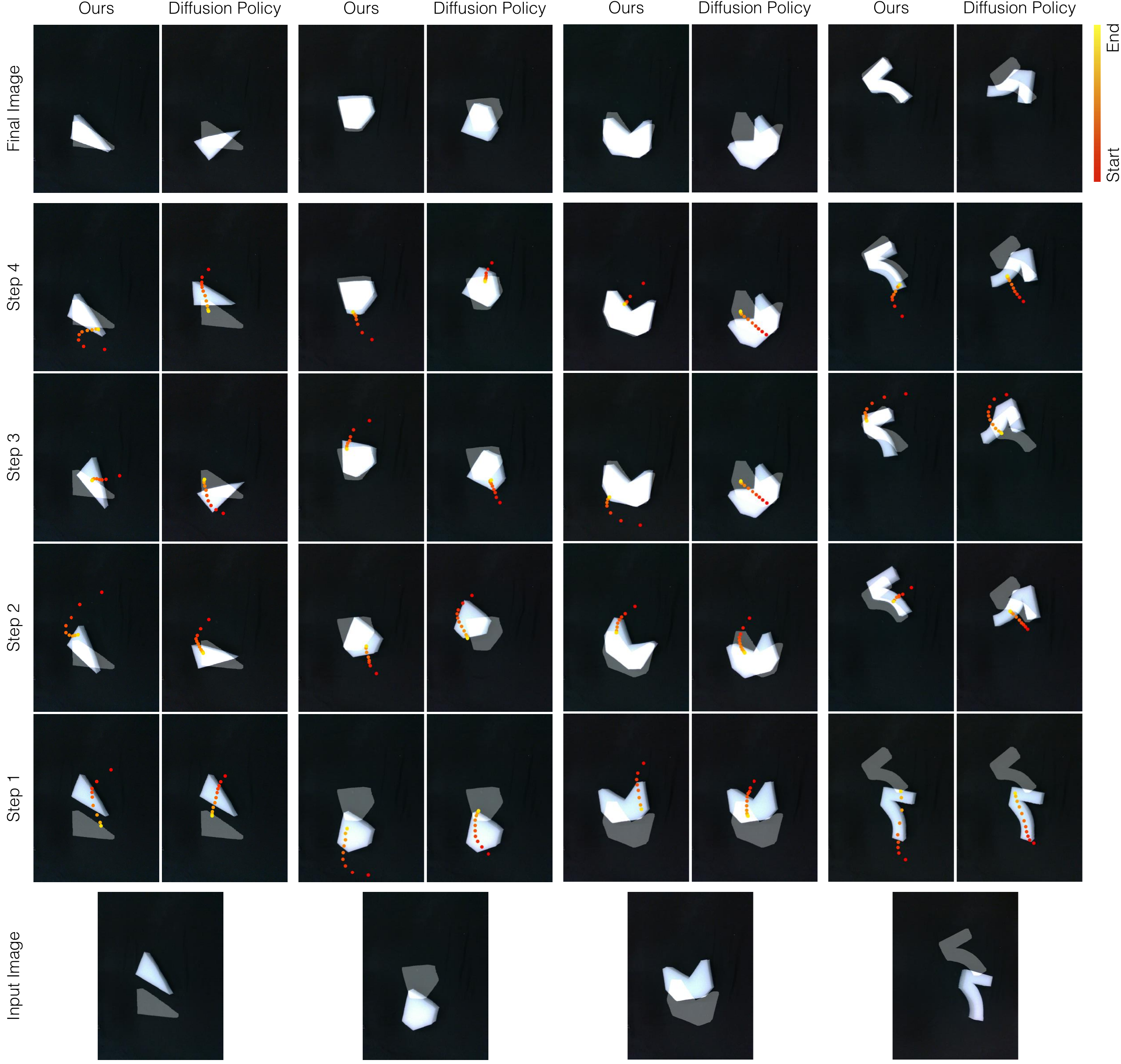}
\caption{\textbf{Additional Push-Shape Qualitative Results.} The trajectory of the end-effector is projected onto the input image.} \label{fig:appendix_push_crop} \end{figure}

We adapted the pretrained Stable Video Diffusion model~\cite{blattmann2023stable}, which generates 25-frame videos at a time. In our adaptation, the first 13 frames correspond to the stereo view 1, and the last 12 frames correspond to the stereo view 2, captured from the two cameras. To condition the video model to generate a stereo video, we modified the per-frame image embedding based on the viewing angle of each output frame. Since each frame of the stereo videos should be paired but the video model generates an odd number of frames, the first frame of the video model output is always the same as the input and discarded at test time. The model training hyperparameters are given in Table~\ref{tab:hyperparameters}. During inference, we use 30 denoising steps with a constant classifier-free guidance of 1.0. Additional qualitative results for the four tasks are shown in Fig~\ref{fig:appendix_rotate_crop}, ~\ref{fig:appendix_scoop_crop}, ~\ref{fig:appendix_sweep_crop} and ~\ref{fig:appendix_push_crop}.

\begin{table}[h]
    \centering
    \small
    \begin{tabular}{cccccccc}
        \toprule
        H-Param & Res & Lr & Batch Size & Train Steps & Clip Duration & Fps & MotScr\\
        \midrule
        Rotation (Full DS) & 768$\times$448 & 1e-5 & 4 & 16384 & 2.0 & 6 & 200 \\
        Rotation (2/3 DS)& 768$\times$448 & 1e-5 & 4 & 16384 & 2.0 & 6 & 200 \\
        Rotation (1/3 DS)& 768$\times$448 & 1e-5 & 3 & 15360 & 2.0 & 6 & 200 \\
        Scooping & 768$\times$448 & 1e-5 & 4 & 16384 & 3.0 & 5 & 200 \\
        Sweeping & 768$\times$448 & 1e-5 & 4 & 16384 & 3.0 & 5  & 200 \\
        Push-Shape & 768$\times$448 & 1e-5 & 4 & 17408 & 2.0 & 6 & 200 \\
        \bottomrule
    \end{tabular}
    \vspace{0.2cm}
    \caption{\textbf{Hyperparameters for Video Model Training.} Res: image and video resolution, Lr: learning rate, Batch Size: batch size, Training Steps: training steps for the evaluation checkpoint, Clip Duration: single demonstration video length in seconds, Fps: the video sub-sampling frame rate and model fps parameter, MotScr: model motion score parameter.}
    \vspace{-1em}
    \label{tab:hyperparameters}
\end{table}

\section{Experimental Setup}

\begin{figure}[h]
\centering
\includegraphics[width=\textwidth]{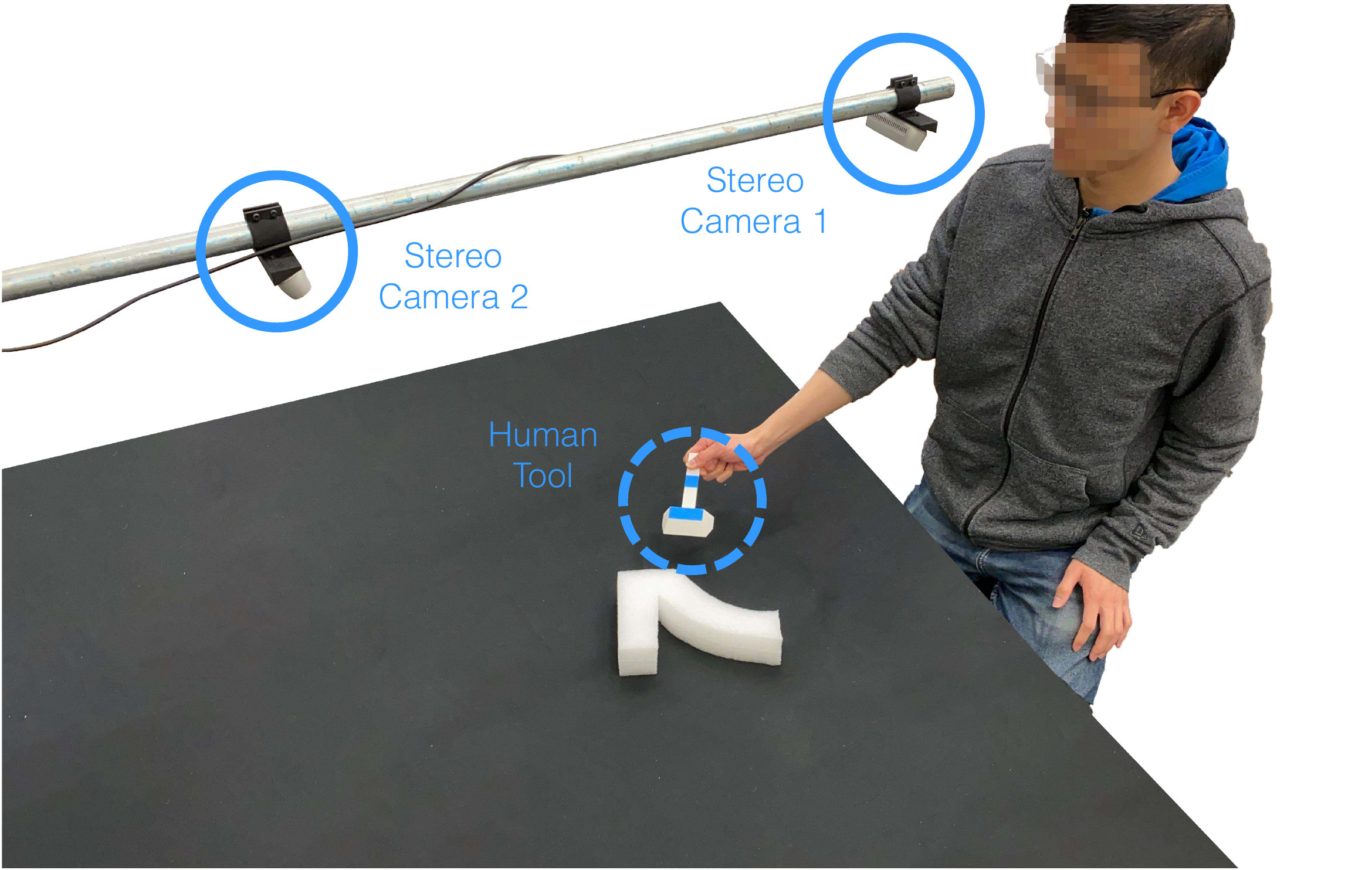}
\caption{\textbf{Real-world Data Collection Setup.} The data collection setup has the same camera arrangement as the robot experiment setup.} \label{fig:human_realworld_setup_cropped} \end{figure}

\begin{figure}[h]
\centering
\includegraphics[width=\textwidth]{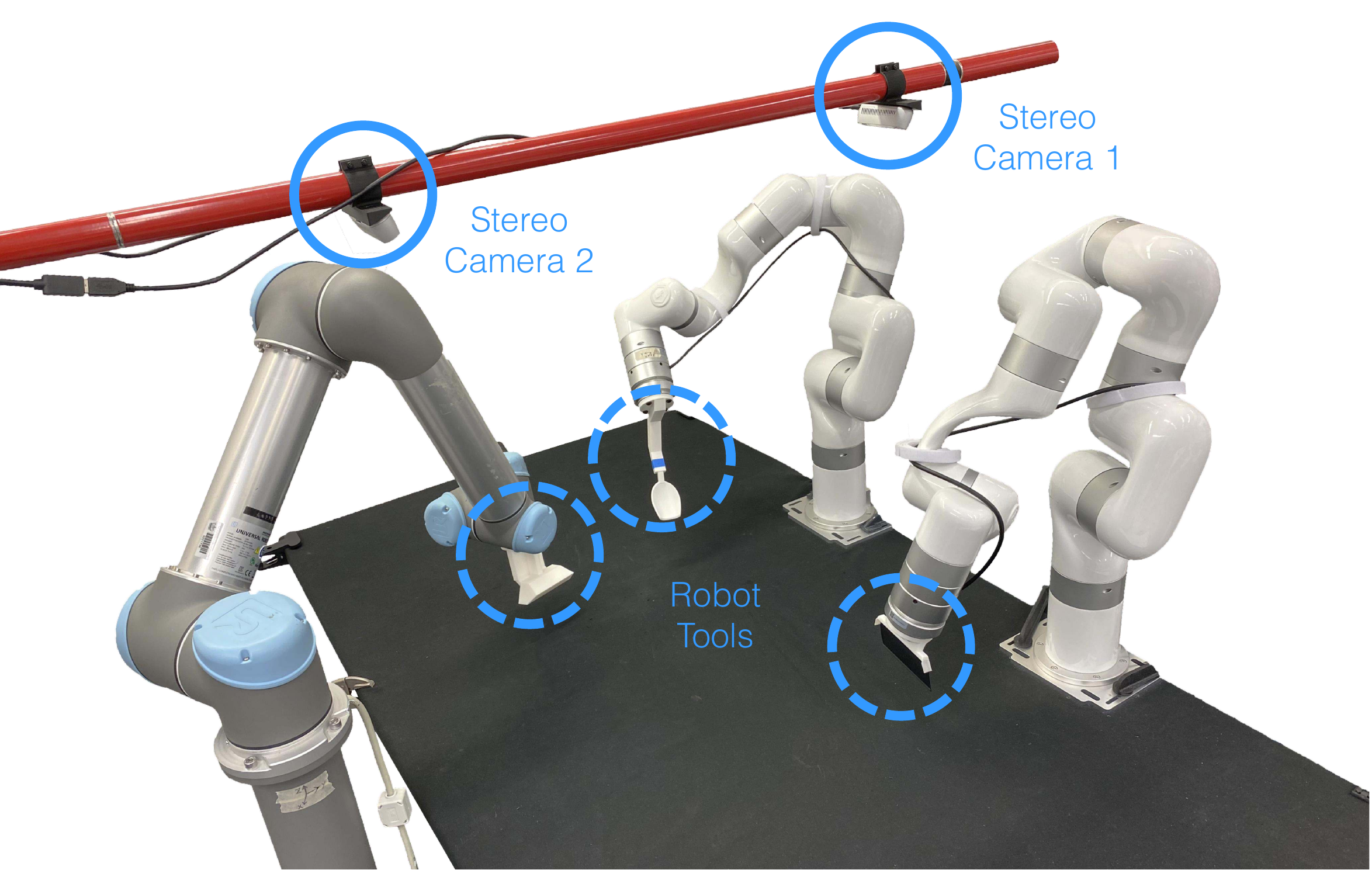}
\caption{\textbf{Real-world Robot Experiment Setup.} The robot experiment setup includes the 3 robots to perform all the experiments.} \label{fig:robot_realworld_setup_cropped} \end{figure}

The stereo camera setup consists of two Intel RealSense D435i cameras spaced approximately 660 mm apart at a 45$^\circ$ angle. The distance between the cameras and the table is about 760 mm. The real-world data collection and the robot experiment setups are shown in Fig.~\ref{fig:human_realworld_setup_cropped} and ~\ref{fig:robot_realworld_setup_cropped}. The training videos are recorded at a resolution of 1280$\times$720 and are then cropped and resized to the appropriate resolution for model input. The table surface used for data collection and experiments is covered with a black cloth, which introduces variations in friction and increases uncertainty in the Push-Shape experiments. For the rotation and scooping experiments, UFACTORY xArm 7 robots are used, while UR5 robots are used for the sweeping and Push-Shape experiments. For calculating the mIoU in the Push-Shape experiment, the view from the stereo camera 1 is used. In each trial with multiple steps, the resulting image with the highest IoU with the target is used to calculate the rotation error. In sweeping and push-shape experiments, the robot end-effector height is limited to avoid robot collision with the table top.

\section{Data Collection}

\begin{figure}[h]
\centering
\includegraphics[width=0.9\textwidth]{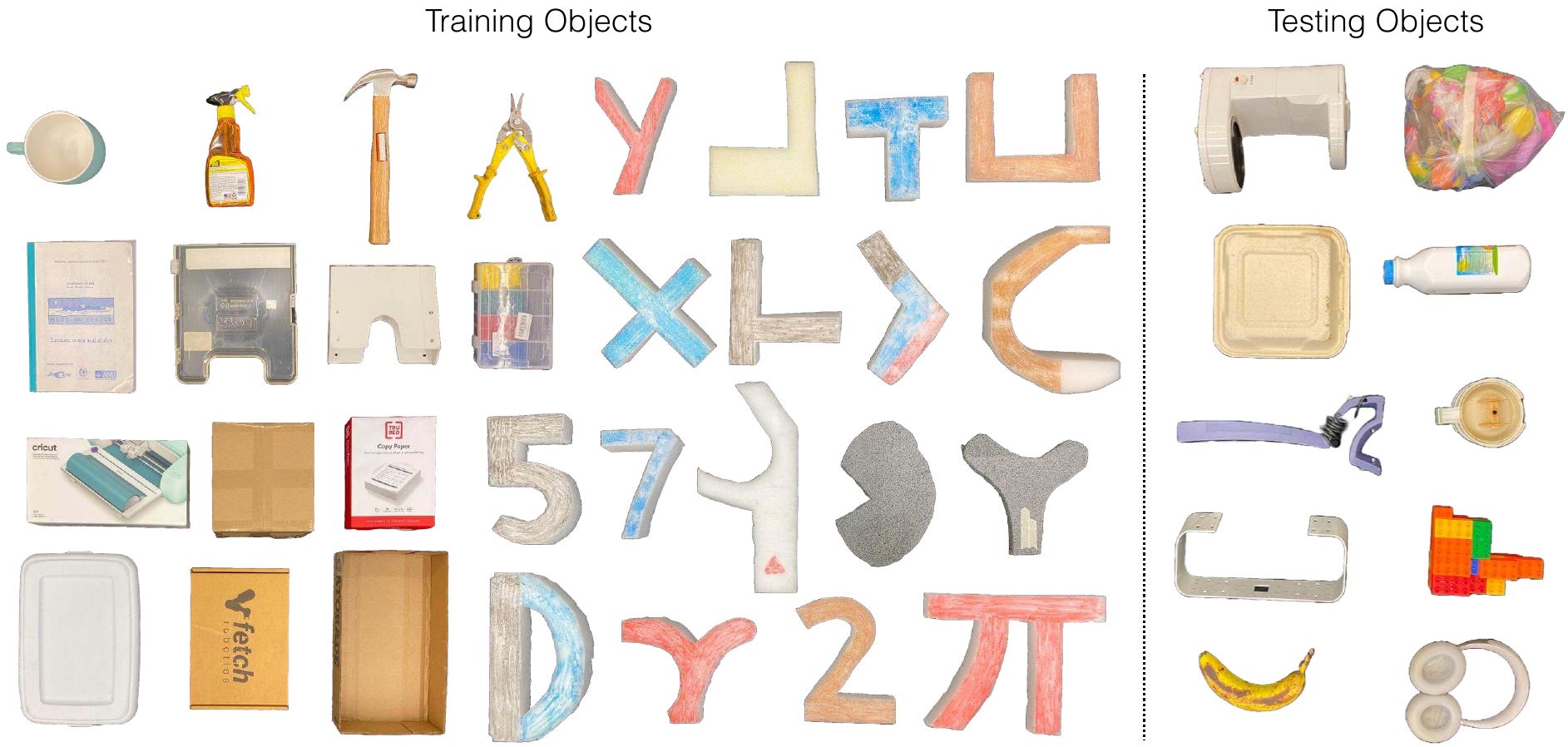}
\caption{\textbf{Rotation Objects.} The training set includes 14 real-world objects and 17 custom colored shapes made out of foam. The testing set includes 10 challenging real-world objects.} \label{fig:rotate_objects} \end{figure}

\begin{figure}[h]
\centering
\includegraphics[width=\textwidth]{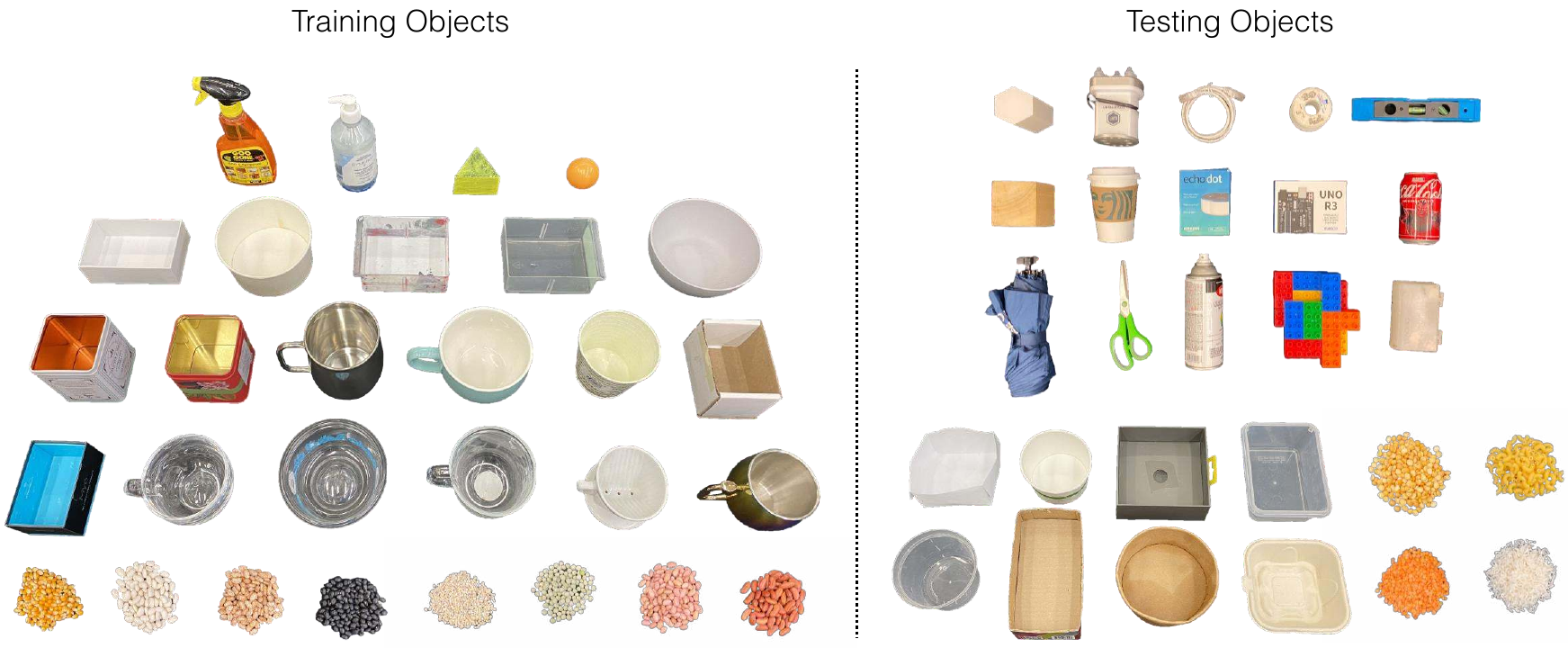}
\caption{\textbf{Scooping Objects.} The training set includes 17 bowls, 8 colored beans and 4 real-world objects. The testing set includes 8 bowls, 4 colored particles and 15 real-world distraction objects. During data collection, both the real-world objects and inverted bowls are used as the single distraction object in the scene.} \label{fig:scoop_objects} \end{figure}

\begin{figure}[h]
\centering
\includegraphics[width=\textwidth]{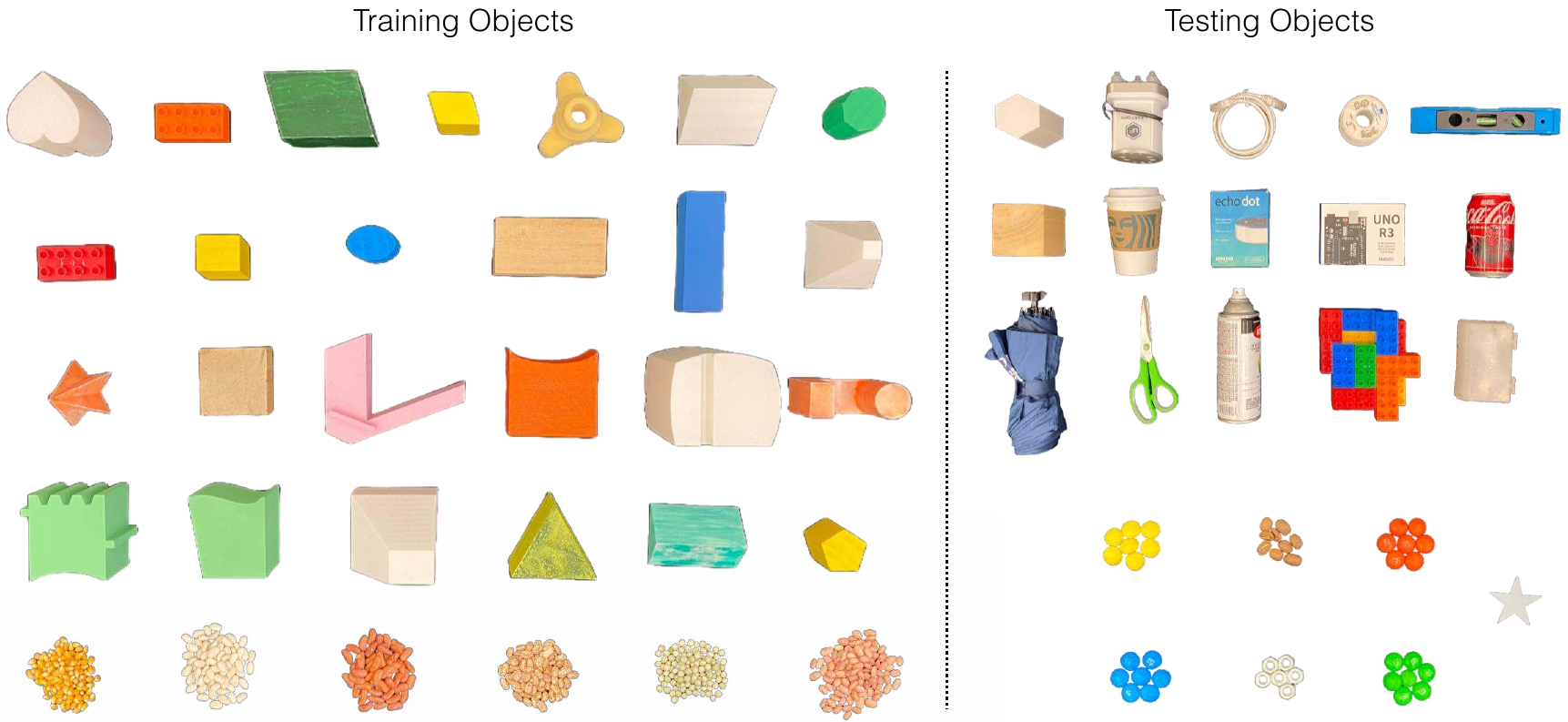}
\caption{\textbf{Sweeping Objects.} The training set includes 25 distraction objects and 6 colored beans, with 5 to 6 distractions in the scene at a time during data collection. The testing set includes 15 real-world distraction objects and 6 colored particles. The star is used as the target in the experiments.} \label{fig:sweep_objects} \end{figure}

\begin{figure}[h]
\centering
\includegraphics[width=\textwidth]{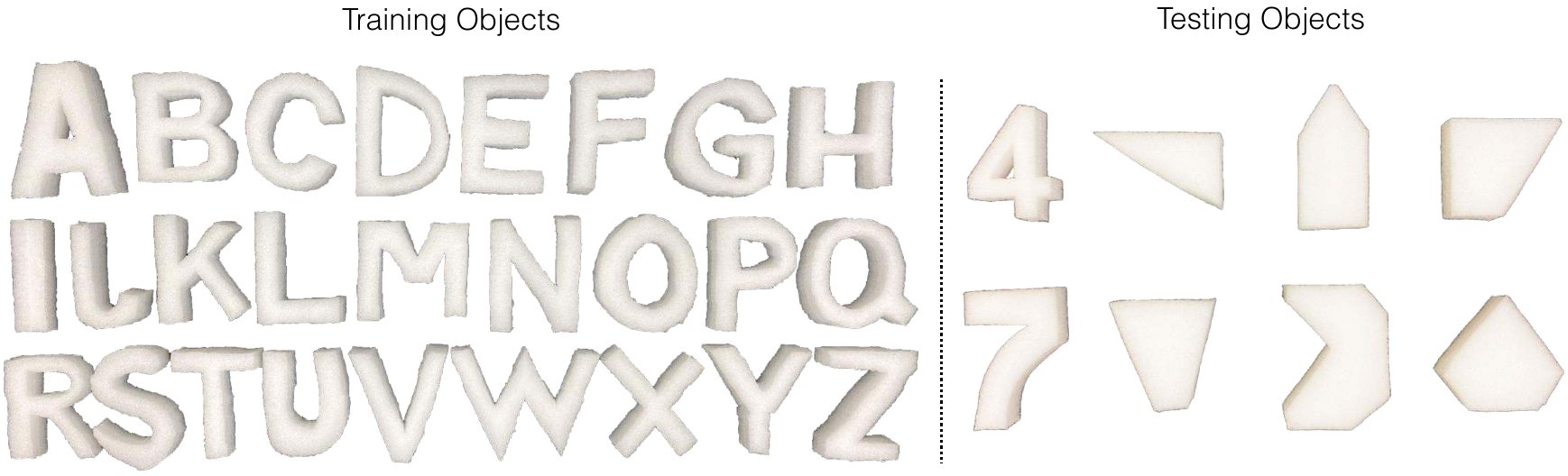}
\caption{\textbf{Push-Shape Objects.} The training set includes all 26 capital letters of the alphabet, while the testing set consists of 8 shapes, including digits and polygons.} \label{fig:push_objects} \end{figure}

For all tasks, the first frame of the human demonstration video is an image of the scene. The subsequent frames include the human demonstrator using the tool to perform the manipulation. In the Push-Shape demonstration, an object is pushed to a location in multiple steps. The final position of the object is used as a mask and blended with the entire video for the target position. The objects used in training and testing for different tasks are shown in Fig.~\ref{fig:rotate_objects},~\ref{fig:scoop_objects},~\ref{fig:sweep_objects} and ~\ref{fig:push_objects}.

\section{Object Tracking}

In the videos, the tool is tracked using MegaPose~\cite{labbe2022megapose}. Utilizing a stereo setup, the center of the tracked object from each camera are projected into 3D space as a straight line. The translation component of the object in 3D space is determined by finding the midpoint between the projected lines from the two cameras. The rotation component of the object is obtained by averaging the object rotations from the two views. This refined object pose from the stereo setup enhances the accuracy of the object's depth measurement from the cameras. In the scooping task, only the handle of the scooper is tracked to avoid inaccuracies due to occlusion by particles. In the sweeping and Push-Shape tasks, the tool without the handle is tracked, as the handle is occluded by the human hand. To obtain the tool trajectories for training Diffusion Policy, the same stereo tracking is applied to the demonstration videos.

\section{Diffusion Policy Baseline}

We use a CNN-based Diffusion Policy as our baseline, employing two pretrained ResNet-18~\cite{he2016deep} image encoders to process the stereo images of the scene. The input images have a resolution of 384$\times$224, similar to the original implementation resolution. We found that higher resolution input images did not improve model performance.


\end{document}